\documentclass{article}
\usepackage{arxiv}
\usepackage[T1]{fontenc}
\usepackage[utf8]{inputenc}
\usepackage[numbers,sort&compress]{natbib}
\usepackage{xspace}
\usepackage{amsmath,amsfonts,amsthm}
\usepackage{algorithm,algorithmic}
\usepackage{graphicx,subfigure}
\usepackage{booktabs,tabularx}
\usepackage{microtype}
\usepackage{flafter}
\usepackage{hyperref}
\usepackage{cleveref}

\DeclareMathOperator{\Orth}{Orth}
\newcommand{\algn}{\ensuremath{\mathsf{DisMUSE}}}
\newcommand{\alg}{\algn\xspace}
\newcommand{\algnns}{\ensuremath{\mathsf{DisMUSE\text{-}NS}}}
\newcommand{\algnlr}{\ensuremath{\mathsf{DisMUSE\text{-}LR}}}
\newcommand{\alglr}{\algnlr\xspace}
\newcommand{\ul}{\underline}
\newcommand{\F}{{\rm F}}

\newtheorem{assumption}{Assumption}
\newtheorem{remark}{Remark}
\newtheorem{theorem}{Theorem}
\newtheorem{lemma}{Lemma}
\newtheorem{corollary}{Corollary}
\theoremstyle{definition}
\newtheorem{definition}{Definition}

\hypersetup{hidelinks,
  pdftitle={DisMUSE: An Efficient Distributed Singular-Spectrum-Based Optimization Approach with Theoretical Convergence Guarantee},
  pdfauthor={Zhen Qin, Zhiyao Zhang, Yingbin Liang, Jia Liu}
}
\title{DisMUSE: An Efficient Distributed\\
Singular-Spectrum-Based Optimization Approach\\
with Theoretical Convergence Guarantee}
\author{Zhen Qin \quad Zhiyao Zhang \quad Yingbin Liang \quad Jia Liu\\
\normalfont Department of Electrical and Computer Engineering\\
\normalfont The Ohio State University, Columbus, Ohio, USA}
\date{}
\renewcommand{\shorttitle}{DisMUSE: Distributed Spectral Optimization}
\begin{document}
\maketitle
\begin{abstract}
In recent years, singular-spectrum-based optimizers have attracted increasing attention due to their ability to preserve the matrix structure of neural network gradients, thereby enabling geometry-aware optimization.
However, most singular-spectrum-based (SSB) approaches remain centralized, limiting their use in distributed learning infrastructures that exploit massive parallelism.
Distributed SSB optimization is still at an early stage, with open questions about convergence guarantees and communication efficiency.
To fill this gap, we propose the \ul{Dis}tributed \ul{M}oment\ul{u}m-based \ul{S}pectral \ul{E}stimator (\algn), an efficient distributed SSB optimization framework for both data-center and federated settings.
Using Muon-style exact orthogonalized momentum as the spectral update, \algn\ integrates a recursive variance-reduced momentum estimator and achieves an $O((KT)^{-1/3})$ convergence rate, where $K$ and $T$ denote the number of nodes and local iterations per node, respectively. 
This rate matches the best-known rates for variance-reduced nonconvex distributed optimization.
Building on this variance-reduced framework, we study two extensions.
First, our \algn\ method with finite Newton-Schulz (NS) iterations preserves the same convergence order under a uniform normalized support-spectrum condition, up to a multiplicative factor.
Second, we develop a communication-efficient variant based on low-rank (LR) momentum transmission and quantify the error it introduces to the stationarity bound, characterizing the trade-off between accuracy and communication cost.
\end{abstract}

\allowdisplaybreaks
\section{Introduction}
Distributed learning has become a core paradigm for modern machine learning systems, spanning data-parallel training, federated learning, and edge learning scenarios where computation and data are spread across multiple nodes.
In such settings, optimization algorithms must handle stochastic gradient noise, statistical heterogeneity, and communication constraints.
While a large body of prior work studies these issues for conventional vector-based methods such as SGD \cite{bottou2010sgd} and Adam \cite{kingma2014adam}, these methods treat parameters as flat vectors.
Much less is understood when parameters and gradients are naturally matrix-valued, as in Transformers and convolutional networks.

Recently, Singular-Spectrum-Based Optimization (SSBO) methods, such as Muon \cite{jordan2024muon}, Shampoo \cite{gupta2018shampoo}, and SOAP \cite{vyas2025soap}, have attracted growing attention for their ability to preserve the matrix structure of neural network updates.
Among them, Muon has garnered the most interest due to its exceptional efficiency in large language model (LLM) training.
Muon also has a relatively simple implementation: it orthogonalizes a matrix-valued momentum before applying the update, yielding geometry-aware descent directions~\citep{liu2025muon,huang2025limuon}. 
Despite the promise of these SSBO methods, both their theory and implementation remain largely confined to centralized training in the literature, leaving open the question of how they should be designed and analyzed in distributed settings.

However, developing SSBO-based methods for distributed learning is highly non-trivial due to several technical challenges.
\textbf{First}, the {\em nonlinear spectral mapping} does not commute with distributed aggregation.
Averaging locally transformed spectral directions is generally not equivalent to applying the spectral map after aggregating the underlying momentum estimates, which can introduce systematic bias.  
\textbf{Second}, data heterogeneity and noisy gradient estimates further exacerbate this nonlinearity challenge.
In distributed stochastic training, deviations in the momentum estimators are coupled with the nonlinear spectral mapping and can lead to directional errors, suggesting that the matrix-valued estimator should be carefully controlled before applying the spectral map.  
\textbf{Third}, practical Muon-style implementations typically replace exact orthogonalization with finite-step Newton–Schulz (NS) iterations. The resulting approximation error must therefore be explicitly accounted for in the convergence analysis.  
\textbf{Fourth}, synchronizing full matrix-valued momentum variables incurs substantial communication cost. This high communication cost motivates low-rank communication, whose additional approximation error must also be carefully controlled.

In this work, we focus on Muon-type SSBO algorithmic framework and propose the \emph{Distributed Momentum-based Spectral Estimator} (\algn) method, a unified framework for distributed SSBO.
The key idea of \alg is to build a variance-reduced distributed SSBO backbone as the foundation for practical Newton-Schulz orthogonalization and communication-efficient low-rank synchronization.
Our contributions are summarized as follows:

\begin{list}{\labelitemi}{\leftmargin=1.3em \itemindent=-.1em  \itemsep=.2em}

\vspace{-.1in}
\item \textbf{A variance-reduced Distributed SSBO Backbone:}
We first develop \alg with exact orthogonalization and full communication of matrix-valued momentum.
A STORM-style recursive estimator controls the stochastic deviation of the matrix-valued momentum before the nonlinear spectral transformation.
Under the stated assumptions, \alg achieves an $O((KT)^{-1/3})$ convergence rate for averaged stationarity, where \(K\) and \(T\) denote the number of nodes and local optimization iterations per node, respectively, matching the best-known rates for variance-reduced nonconvex distributed optimization.

\item \textbf{\alg with Finite-step Newton-Schulz Orthogonalization:}
We replace the exact orthogonalization step with a finite number of Newton-Schulz (NS) iterations and show that the same $O((KT)^{-1/3})$ convergence order is preserved under a uniform normalized support-spectrum condition, up to a multiplicative approximation factor.

\item \textbf{Communication-efficient Low-rank Implementation:}
We extend \alg to a low-rank communication-efficient implementation, where nodes can transmit low-rank momentum factors instead of full matrices. 
We quantify the additional error introduced by low-rank communication through an additive accumulated term accounting for both client and server spectral truncation.
\end{list}

\section{Related Work}

In this section, we provide an overview on three closely related research areas to put our work into comparative perspectives.

\smallskip
\textbf{1) Singular-Spectrum-Based Optimization and Muon:}
Recent progress in large-scale neural network training has led to a class of optimization methods that explicitly exploit the matrix structure of model updates by manipulating their singular spectra, rather than applying element-wise preconditioning, such as Shampoo \cite{gupta2018shampoo}, SOAP \cite{vyas2025soap}, and Muon \cite{jordan2024muon}.
These methods have demonstrated strong empirical performance on a variety of deep learning tasks \citep{ahn2025dion,anil2020scalable,jordan2024muon,liu2025muon,petrov2025signmuon,shah2025practical,wang2025muon,zhang2025adagrad}.
In the literature, these methods are sometimes loosely referred to as ``matrix-aware optimization'' or ``geometric optimization.''
However, ``matrix-aware methods'' may respect tensor structure without necessarily operating on spectral properties (e.g., Adafactor \citep{shazeer2018adafactor}), while ``geometric optimization'' broadly covers manifold-based updates without isolating spectrum-level control \citep{absil2008optimization}.
To avoid confusion, in this paper, we use the \emph{Singular-Spectrum-Based Optimization (SSBO)} terminology for methods that directly manipulate the singular spectrum of matrix updates.

Among SSBO-based approaches, Muon has garnered increasing attention.
Muon orthogonalizes a momentum matrix before applying the update, thereby aligning the descent direction with the spectral geometry of the matrix-valued gradient \citep{bernstein2024old}.
In practice, Muon typically avoids exact singular value decomposition (SVD) by using iterative approximations such as NS orthogonalization \citep{jordan2024muon}.
Recent work has studied the convergence properties, geometric interpretations, and optimization benefits of Muon and related spectral optimizers \citep{ahn2025dion,chang2025convergencemuon,huang2025limuon,anonymous2025convergence,kovalev2025trust,lau2025polargrad,li2025note,liu2025cosmos,mahdavinia2025lowrank,pethick2025normconstrained,sato2025batchsize,shen2025convergence,si2025adamuon}.
However, existing studies remain largely focused on centralized training, leaving distributed SSBO with practical NS orthogonalization less understood.

\smallskip
\textbf{2) Variance Reduction and Communication-Efficient Distributed Learning:}
Distributed learning in data-center, federated, and decentralized settings must address the challenges of communication costs, stochastic noise, and data heterogeneity \citep{karimireddy2020scaffold,mcmahan2017communication,nedic2009distributed,reddi2020adaptive,stich2018local}.
Communication efficiency is commonly addressed through local updates \citep{mcmahan2017communication,stich2018local} and compression techniques such as quantization, sparsification, and low-rank approximation \citep{aji2017sparse,alistarh2017qsgd,basu2019qsparse,bernstein2018signsgd,haddadpour2021federated,reisizadeh2020fedpaq,stich2018sparsified,vogels2019powersgd,wang2018atomo}, while momentum-based variance-reduction schemes address node drift and stochastic noise under heterogeneous data distributions \citep{khanduri2021stem}.
Decentralized learning further relies on peer-to-peer communication, with methods based on model averaging, compressed communication, and Gradient Tracking \citep{koloskova2019decentralized,lian2017can,nedic2009distributed,nedic2017achieving,pu2021distributed}.
However, these methods are developed for conventional vector-based optimization and do not explicitly account for nonlinear spectral transformations applied to matrix-valued momentum or gradient estimators.

\smallskip
\textbf{3) Muon in Federated and Distributed Settings:}
Recent works have started to investigate Muon-style optimization in the FL setting.
Existing studies have considered the theoretical benefits of Muon in FL, including communication complexity, robustness to heavy-tailed noise, and learning-rate choices \citep{zhang2025provable}; bias-corrected federated Muon with control variates and inexact NS iterations \citep{takezawa2026fedmuon}; and local-global alignment to mitigate client drift under heterogeneity \citep{liu2025fedmuon}.
For convenience, we refer to these baseline methods as FedMuon-FA \citep{zhang2025provable}, FedMuon-SF \citep{takezawa2026fedmuon}, and FedMuon-LRSF \citep{liu2025fedmuon}, respectively.
Compared to these federated Muon methods, our main goal is to build a \emph{variance-reduced distributed SSBO backbone}.
\algn\ controls the stochastic deviation of the matrix-valued momentum estimator before applying the nonlinear spectral mapping, and uses this backbone for a unified treatment of exact orthogonalization, finite-step NS approximation, and communication-efficient low-rank synchronization.
Such variance-reduced mechanisms were not considered in the above FedMuon-type methods \citep{liu2025fedmuon,takezawa2026fedmuon,zhang2025provable}.

\section{Problem Statement and Preliminaries}
We first state the distributed matrix-valued optimization problem, followed by spectral mapping and exact orthogonalization. We then discuss the key challenges in distributed SSBO that motivate \algn{}.

\subsection{Distributed Matrix-Valued Optimization} \label{subsec:problem_statement}
We consider distributed optimization over a matrix-valued parameter $W\in\mathbb{R}^{m\times n}$ with $K$ nodes indexed by $k\in\{1,\ldots,K\}$.
Each node has access to a local data distribution $\mathcal D_k$ and defines
$
f^{(k)}(W)=\mathbb E_{\xi_k\sim \mathcal D_k}[f^{(k)}(W;\xi_k)],
$
where $f^{(k)}(W;\xi)$ denotes the sample-wise loss.
The global objective is
\[
\min_{W\in\mathbb R^{m\times n}} f(W):=\frac1K\sum_{k=1}^K f^{(k)}(W).
\]
We consider the nonconvex setting, where each $f^{(k)}$ is nonconvex in $W$, and assume that the global objective is bounded below, with \(f_*:=\inf_W f(W)>-\infty\). Our goal is to find an approximate stationary solution.
We use upper-case letters for matrices.
For $A\in\mathbb R^{m\times n}$, let $\{\sigma_i(A)\}_{i=1}^{\min(m,n)}$ denote its singular values, and let $\|A\|_F$, $\|A\|_{\mathrm{op}}$, and $\|A\|_*$ denote the Frobenius, spectral, and nuclear norms, respectively: $\|A\|_F=(\sum_i\sigma_i(A)^2)^{1/2}$, $\|A\|_{\mathrm{op}}=\max_i\sigma_i(A)$, and $\|A\|_*=\sum_i\sigma_i(A)$.
They satisfy $\|A\|_{\mathrm{op}}\le \|A\|_F\le \|A\|_*$, and, if $\operatorname{rank}(A)\le r$, $\|A\|_*\le \sqrt r\,\|A\|_F$.

\subsection{Spectral Mapping and Orthogonalization for SSBO-Based Optimization} \label{subsec:spectral_map}

In matrix-valued optimization, SSBO applies a nonlinear spectral mapping to a matrix-valued gradient or momentum estimator.
In Muon-style methods, this mapping is an exact orthogonalization.
For a matrix $M \in \mathbb{R}^{m \times n}$ with singular value decomposition $M = U\Sigma V^\top$, we define
$
\mathrm{Orth}(M) := U_+ V_+^\top,
$
where $U_+$ and $V_+$ contain the singular vectors corresponding to positive singular values, and $\mathrm{Orth}(0)=0$.
Muon applies this mapping to a momentum matrix $M_t$.
In the centralized setting, an orthogonalized Muon update on the matrix weight $M_t$ can be written as:
\[
\begin{aligned}
M_{t+1} &= \beta M_t + (1-\beta)\nabla f(W_t;\xi_t), \\
W_{t+1} &= W_t - \eta_t \mathrm{Orth}(M_{t+1}).
\end{aligned}
\]
The key challenge in distributed SSBO is that $\mathrm{Orth}(\cdot)$ is nonlinear, resulting in key challenges in distributed SSBO-based algorithmic design as stated next.

\subsection{Key Challenges in Distributed SSBO-based Algorithmic Design} \label{subsec:challenges}
Although SSBO-based approaches have shown promise in centralized training, developing distributed SSBO-based algorithms presents several structural challenges that do not arise in their centralized counterparts.
These challenges mainly arise from the nonlinear spectral mapping applied to matrix-valued gradient or momentum estimators, which interacts with data heterogeneity, stochastic noise, and communication constraints.
In what follows, we discuss these issues in detail and use them to motivate the design of \algn.

\smallskip
\textbf{Challenge 1) Non-Commutativity and Aggregation Bias:}
A first difficulty comes from the fact that the nonlinear spectral mapping does {\em not} commute with distributed aggregation.
Let $M_t^{(k)}$ denote the local momentum estimator maintained by node $k$ at iteration $t$.
Then, we have:
\[
\frac{1}{K}\sum_{k=1}^K \mathrm{Orth}(M_t^{(k)})
\neq
\mathrm{Orth} \bigg(\frac{1}{K}\sum_{k=1}^K M_t^{(k)}\bigg).
\]
As a result, averaging locally orthogonalized momentum directions is {\em not} equivalent to  the spectral direction of the averaged global momentum in the centralized setting. 
This differs fundamentally from vector-based distributed optimization, where commutativity holds due to the linear averaging operation.
The resulting discrepancy can introduce systematic bias, especially when local objectives or stochastic gradients differ across nodes.
The non-commutativity suggests a key design principle in \algn: {\em the algorithm should synchronize and control the matrix-valued momentum estimators before averaging locally transformed spectral directions.}

\smallskip
\textbf{Challenge 2) Variance Amplification under Nonlinear Spectral Mapping:}
A second challenge arises from stochastic noise before applying the nonlinear spectral mapping.
Let
$
M_t = M_t^{\mathrm{det}} + E_t, \mathbb{E}[E_t]=0,
$
where $M_t^{\mathrm{det}}$ is the deterministic momentum and $E_t$ is a stochastic estimation error.
In conventional vector-based updates, $E_t$ emerges in the update in an additive fashion.
In SSBO, however, the corresponding update perturbation becomes
\[
\Orth(M_t^{\mathrm{det}}+E_t)
-
\Orth(M_t^{\mathrm{det}}),
\]
which is not a linear or additive function of $E_t$.
As a result, even if $\mathbb{E}[E_t]=0$, one still has $\mathbb{E}[\Orth(M_t)] \neq \Orth(M_t^{\mathrm{det}})$.
Thus, stochastic noise in the momentum estimator can lead to biased and directionally distorted spectral updates.
This makes variance control important not only for efficiency, but also for obtaining robust spectral directions.
\alg addresses this issue using a STORM-style recursive estimator \cite{cutkosky2019momentum} to reduce the stochastic deviation of the matrix-valued momentum before applying the spectral mapping.

\smallskip
\textbf{Challenge 3) Orthogonalization Approximation Error:}
Our basic version of \alg is based on exact orthogonalization.
However, in practical Muon-style implementations, exact orthogonalization is typically approximated by a finite number of low-complexity Newton--Schulz (NS) iterations.
Let $X_{t,q}$ denote the result after $q$ NS iterations applied to $M_t$.
Then the actual update direction $X_{t,q}$ differs from $\Orth(M_t)$ by the approximation error $X_{t,q}-\Orth(M_t)$, which may not be negligible.
Moreover, this approximation error resides within the nonlinear spectral transformation and may propagate across iterations.
This motivates our NS extension, where we theoretically quantify the finite-step approximation error and show that \alg retains the same finite-time convergence order under a uniform normalized support-spectrum condition.

\smallskip
\textbf{Challenge 4) Communication Constraints:}
Lastly, distributed SSBO must account for the communication cost of synchronizing matrix-valued momentum variables.
For $M\in\mathbb{R}^{m\times n}$, full communication requires transmitting $mn$ scalars per synchronization, which can be prohibitive in federated and bandwidth-constrained settings.
A natural approach is to exploit low-rank structure and transmit low-rank momentum approximations.
However, low-rank factorization is itself nonlinear and can introduce additional bias through truncation and aggregation.
This motivates our low-rank extension of \algn, where the resulting approximation error is explicitly controlled.

\section{\algn: Variance-reduced Distributed SSBO with Exact Orthogonalization}\label{sec:fed1}\label{dmf:sec:exact}
In this section, we start with the core algorithmic design of the basic version \alg with exact orthogonalization (through exact SVD) and full-rank communication.
The key idea is to synchronize the matrix-valued momentum estimator across nodes and to control its stochastic variance before applying the nonlinear spectral map.

\begin{algorithm}[!t]
\caption{Unified \algn{} Framework}
\label{alg:fedmuon_unified}
\begin{algorithmic}[1]
\STATE \textbf{Input:} stepsize $\eta>0$, $\beta\in(0,1)$, integer period $\tau\ge1$, initial batch size $B_0$, horizon $T$, rank $r\in\{1,\ldots,d\}$; exact or NS orthogonalization (NS: $q\ge0$, $\kappa\ge1$, scale rule in Section~\ref{dmx:sec:ns})
\STATE Initialize $W_0^{(k)}=W_0$ for all nodes
\STATE Form and synchronize the dense $M_0^{(k)}=\bar M_0$ by \eqref{dmf:eq:init}
\FOR{$t=0,\ldots,T-1$, \textbf{on each node} $k$}
    \STATE Compute $O_t^{(k)}$ by the selected branch of \eqref{dmf:eq:local-model}
    \STATE $W_{t+1}^{(k)}\gets W_t^{(k)}-\eta O_t^{(k)}$
    \IF{$(t+1)\bmod\tau=0$}
        \STATE Average and broadcast dense models: $W_{t+1}^{(k)}\gets K^{-1}\sum_jW_{t+1}^{(j)}$
    \ENDIF
    \STATE Draw a fresh sample $\xi_{t+1}^{(k)}$
    \STATE Compute $\widetilde M_{t+1}^{(k)}$ by \eqref{dmf:eq:storm}, using the same sample at both model positions
    \IF{$(t+1)\bmod\tau=0$}
        \STATE Set $s=t+1$
        \IF{$r=d$}
            \STATE Average and broadcast dense momentum: $M_s^{(k)}\gets K^{-1}\sum_j\widetilde M_s^{(j)}$
        \ELSE
            \STATE Client $k$ computes a truncated SVD:\newline $M_{s,r}^{(k)}=U_{s,r}^{(k)}\Sigma_{s,r}^{(k)}(V_{s,r}^{(k)})^\top \approx\widetilde M_s^{(k)}$
            \STATE Upload only $(U_{s,r}^{(k)},\Sigma_{s,r}^{(k)},V_{s,r}^{(k)})$ (diagonal entries of $\Sigma$ only)
            \STATE Server reconstructs and averages matrices:\newline $\bar M_{s,r}\gets K^{-1}\sum_j U_{s,r}^{(j)}\Sigma_{s,r}^{(j)}(V_{s,r}^{(j)})^\top$
            \STATE Server recompresses by truncated SVD:\newline $M_{s,r}=U_{s,r}\Sigma_{s,r}V_{s,r}^\top\approx\bar M_{s,r}$
            \STATE Broadcast only $(U_{s,r},\Sigma_{s,r},V_{s,r})$; all nodes reconstruct $M_s^{(k)}\gets M_{s,r}$
        \ENDIF
    \ELSE
        \STATE $M_{t+1}^{(k)}\gets\widetilde M_{t+1}^{(k)}$
    \ENDIF
\ENDFOR
\end{algorithmic}
\end{algorithm}

\smallskip
{\textbf{1) Algorithm Description:}
We consider a distributed protocol with $K$ nodes and communication period $\tau$. Each node $k$ maintains a local model $W_t^{(k)}$ and a matrix-valued momentum estimator $M_t^{(k)}$. In the $t$-th communication round, the nodes synchronize both model parameters and momentum estimators as follows: $W_t^{(k)}\leftarrow\frac1K\sum_{k'=1}^K W_t^{(k')}$, $M_t^{(k)}\leftarrow\frac1K\sum_{k'=1}^K M_t^{(k')}$. We note that synchronizing $M_t^{(k)}$ first is important because the spectral map is nonlinear, which implies non-commutativity: applying $\Orth(\cdot)$ after momentum aggregation is different from averaging locally orthogonalized directions in general.

Between communication rounds, each node draws a fresh sample $\xi_{t+1}^{(k)}$ and forms the STORM-style estimator \cite{cutkosky2019momentum}
\begin{align}
\widetilde M_{t+1}^{(k)}={}&\beta M_t^{(k)}+(1-\beta)\nabla f^{(k)}(W_{t+1}^{(k)};\xi_{t+1}^{(k)})\label{dmf:eq:storm}\\
&+\beta\bigl(\nabla f^{(k)}(W_{t+1}^{(k)};\xi_{t+1}^{(k)})-\nabla f^{(k)}(W_t^{(k)};\xi_{t+1}^{(k)})\bigr).\notag
\end{align}
The model is then updated using exact orthogonalization as follows: 
$$W_{t+1}^{(k)}=W_t^{(k)}-\eta\Orth(M_t^{(k)}).$$
The recursive correction term above reduces the stochastic deviation of $M_{t}^{(k)}$, which is essential before applying the nonlinear spectral mapping. The full algorithm is illustrated in Algorithm~\ref{alg:fedmuon_unified}.

\smallskip
{\textbf{2) Convergence Analysis}:}
Before conducting convergence analysis, we first introduce several useful assumptions that are standard in nonconvex distributed stochastic optimization. 
\begin{assumption}[Smoothness]\label{ass: smooth}\label{dmf:ass:smooth}
For each node $k$ and any $W,W' \in \mathbb{R}^{m\times n}$,
$
\mathbb{E}_{\xi\sim\mathcal{D}_k}
[
\|
\nabla f^{(k)}(W;\xi)
-
\nabla f^{(k)}(W';\xi)
\|_F^2
]
\le
L^2\|W-W'\|_F^2.
$
\end{assumption}
By Jensen's inequality, Assumption~\ref{ass: smooth} also implies that each local objective $f^{(k)}$ is $L$-smooth with respect to the Frobenius norm, extending the standard Lipschitz-smooth condition from vector-valued to matrix-valued optimization problems~\citep{shen2025convergence}.

To facilitate stochastic optimization convergence analysis, we use the following standard bounded variance assumption:
\begin{assumption}[Bounded variance]\label{ass: variance}\label{dmf:ass:variance}
For each node $k$, the stochastic gradient is unbiased and has bounded variance: $\mathbb{E}[\nabla f^{(k)}(W;\xi_k)] = \nabla f^{(k)}(W)$ and $\mathbb{E}[\|\nabla f^{(k)}(W;\xi_k)-\nabla f^{(k)}(W)\|_F^2]\le \sigma^2$.
All fresh samples are drawn independently across nodes and iterations, including initialization samples.
\end{assumption}
Further, we assume that the statistical data heterogeneity across nodes is bounded in the sense of gradient Frobenius norm: 
\begin{assumption}[Bounded heterogeneity]\label{assumption:heterogeneity}\label{dmf:ass:heterogeneity}
The local gradients satisfy $\frac{1}{K}\sum_{k=1}^K\mathbb{E}[\|\nabla f^{(k)}(W)-\nabla f(W)\|_F^2]\le \delta^2$, where $\delta \ge 0$ measures the degree of data heterogeneity.
\end{assumption}
Note that when $\delta=0$, the data distributions across nodes are homogeneous.
In this work, we measure the convergence of our \alg algorithm using a Frobenius-norm stationarity metric, which generalizes the conventional Euclidean stationarity criterion to matrix-valued problems.

\begin{definition}[$\epsilon$-Frobenius-norm Stationarity]
A solution $\widehat W$ is an $\epsilon$-Frobenius stationary solution of $f(\cdot)$ if $\|\nabla f(\widehat W)\|_F \le \epsilon.$
\end{definition}

Next, we present the main convergence result of the basic \algn, which is based on exact orthogonalization and full-rank communications.
The analysis shows that once the stochastic variance of the matrix-valued momentum is well controlled before spectral mapping, our \alg method enjoys the {\em same} finite-time convergence rate as variance-reduced nonconvex distributed optimization.
Let
$\bar W_t := \frac{1}{K}\sum_{k=1}^K W_t^{(k)}$ represent the averaged model.
We have:
\begin{theorem}[Convergence of \algn{} with exact orthogonalization]
\label{dmf:thm:main}
Suppose Assumptions~\ref{dmf:ass:smooth}--\ref{dmf:ass:heterogeneity} hold.
For fixed $\tau$, $T\ge\max\{K^2,\tau\},r=d$, and
$\eta=K^{1/3}/T^{2/3}$, $1-\beta=K^{1/3}/(2T^{2/3})$, and
$B_0=\lceil T^{1/3}\rceil$, Algorithm~\ref{alg:fedmuon_unified}
with exact orthogonalization satisfies
\begin{equation}
\frac1T\sum_{t=0}^{T-1}\mathbb E\|\nabla f(\bar W_t)\|_F
\le
\frac{C_\tau}{(KT)^{1/3}},
\label{dmf:eq:main-rate}
\end{equation}
where $C_\tau$ is independent of $K,T$ under the stated conditions.
\end{theorem}

The result in \Cref{dmf:thm:main} says that incorporating recursive variance control in \alg yields an $O((KT)^{-1/3})$ stationarity guarantee under exact orthogonalization. 
The proof is provided in Appendix~\ref{dmf:app:main}. 
Compared with the $O((KT)^{-1/4})$ rate reported for existing federated SSBO methods based on standard momentum (FedMuon-FA \citep{zhang2025provable}, FedMuon-SF \cite{takezawa2026fedmuon} and FedMuon-LRSF \cite{liu2025fedmuon}), our convergence result obtains a faster rate under the additional mean-square stochastic smoothness assumption.
This highlights the importance of variance-reduced estimators when applying nonlinear spectral mappings in distributed settings, where update directions are particularly sensitive to gradient estimation error.

\section{\alg with Newton-Schulz Approximate Orthogonalization}
\label{dmx:sec:ns}\label{sec:fed2}

In the previous section, we have shown the strong convergence performance of the basic \alg method based on exact orthogonalization.
However, computing exact $\Orth(M)$ through SVD-based orthogonalization can be expensive for large matrices.
In practical Muon-style implementations, exact orthogonalization is often replaced by a finite number of low-complexity NS steps.
In this section, we will establish the theoretical convergence guarantees of our \alg method with NS orthogonalization approximation.

\smallskip
{\textbf{1) Preliminaries on the Newton-Schulz Approximation and Algorithmic Updates in \algn:}} %
NS approximates the exact orthogonal factor $\Orth(M_t)$. We suppress the node superscript $(k)$ when describing the NS iteration and its residuals.
Starting from a rescaled matrix of \(M_t\) denoted as \(X_{t,0}=M_t/\alpha_t\), with \(\alpha_t\ge \max\{1,\|M_t\|_{\F}\}\) to ensure \(\|X_{t,0}\|_{\F}\le 1\), NS approximation can be iteratively written as:
\[
X_{t,j+1}=p_\kappa(X_{t,j}X_{t,j}^{\top})X_{t,j},
\qquad j=0,\ldots,q-1,
\]
where \(p_\kappa\) is a polynomial approximation of \(\lambda^{-1/2}\) defined as
$
p_{\kappa}(\lambda)
= \sum_{s=0}^{\kappa} c_{s}(1-\lambda)^{s},
c_{s}=\frac{(2 s)!}{4^{s}(s!)^{2}} > 0 .
$
Intuitively, this iteration pushes the singular values of \(X_{t,j}\) toward \(1\), so \(X_{t,q}\) approaches the exact orthogonal factor \(\Orth(M_t)\).

Due to space limitations and to avoid repetition, we omit the description of \alg with the NS approximation, as it differs from the basic \alg only in the orthogonalization step.
In what follows, we outline how to replace the exact orthogonalization with a finite-step NS approximation:
In $t$-th iteration at node $k$, we first apply $q$ NS steps on $M_t^{(k)}$ to obtain $X^{(k)}_{t,q}$, which serves as an approximation to the exact orthogonal factor.
Then, the update direction is computed as:
\begin{align} \label{eqn:NS_dir}
    O_t^{(k)} = X^{(k)}_{t,q}.
\end{align}

\smallskip
{\textbf{2) Convergence Analysis:}}
To conduct convergence analysis of \alg under NS orthogonalization approximation, we adopt the following assumption:
\begin{assumption}[Spectral norm Lipschitz smoothness]\label{ass: nuclear smooth}
Each local objective $f^{(k)}$ is $L_*$-smooth with respect to the spectral/nuclear norm pair: for any $W,W' \in \mathbb{R}^{m \times n}$,
it holds that $
\|\nabla f^{(k)}(W) - \nabla f^{(k)}(W')\|_*
\le L_*\|W-W'\|_{\mathrm{op}} .
$
\end{assumption}
This assumption captures smoothness properties aligned with the spectral structure of matrix-valued gradients and will be used when measuring stationarity in terms of the nuclear norm \citep{chen2025muon}.
Moreover, for more convenience in analyzing NS steps in spectral approximation, we adopt the following nuclear-norm stationarity.

\begin{definition}[Nuclear-norm stationarity]
A solution $\widehat W$ is said to be an $\epsilon$-nuclear stationary solution of $f(\cdot)$ if $\|\nabla f(\widehat W)\|_* \le \epsilon.$
\end{definition}

Note that, since $\|A\|_F \le \|A\|_*$, nuclear-norm stationarity implies Frobenius-norm stationarity (i.e., our convergence result is stronger than needed for the Frobenius norm), which reconciles with our previous convergence results for the basic \alg method. 

To quantify the NS approximation error after $q$ steps, let $X_{t,q}$ denote the resulting matrix at iteration $t$ and let $\Pi_t$ be the orthogonal projector onto the range of the corresponding momentum matrix $M_t$.
We define orthogonality residual as $\delta_{t,q} := \| \Pi_t - X_{t,q} X_{t,q}^\top \|_{\mathrm{op}}$ and polar approximation error as $\varepsilon_{t,q} := \| X_{t,q} - \mathrm{Orth}(M_t) \|_{\mathrm{op}}.$
These quantities capture the deviation from perfect orthogonality and the error relative to exact orthogonalization, respectively.
Then, we invoke a known quantitative bound on the approximation error induced by the NS iteration, which characterizes how the number of iterations $q$ controls the deviation from exact orthogonalization and will be used in our convergence analysis.

\begin{lemma}[Restated from \cite{anonymous2025convergence}]\label{lem:reuse}\label{dmx:lem:ns}
For the Newton--Schulz polynomial with degree $\kappa$ and any $t$,
$
\delta_{t,q} \le \delta_{t,0}^{(\kappa+1)^q}.
$
Consequently, the polar approximation error and the associated amplification factor satisfy
$\varepsilon_q \le 1-\sqrt{1-\delta_0^{(\kappa+1)^q}} \le \delta_0^{(\kappa+1)^q},$ and $\chi_q := \frac{1}{1-\varepsilon_q}
\le \frac{1}{\sqrt{1-\delta_0^{(\kappa+1)^q}}},$
where $\delta_0 := \sup_t \delta_{t,0} < 1$.
\end{lemma}
This result implies that the approximation error $\varepsilon_q$ decays at a doubly exponential rate in the number of NS steps $q$, and the associated factor $\chi_q$ converges to $1$ at the same rate.
Our main convergence result for the \alg method with NS orthogonalization approximation is stated as follows:
\begin{theorem}[Convergence with finite-step NS]
\label{dmx:thm:ns}
Suppose Assumptions~\ref{dmf:ass:smooth}--\ref{ass: nuclear smooth} hold.
Assume that $\delta_{t,0}^{(k)}\le\delta_0<1$ almost surely
for all nodes and iterations, where $\delta_0$ is deterministic
and independent of $K,T$.
For fixed $\tau$, $T\ge\max\{K^2,\tau\},r=d$, and the parameter choices of Theorem~\ref{dmf:thm:main}, Algorithm~\ref{alg:fedmuon_unified} with finite-step NS satisfies
\begin{equation}
\frac1T\sum_{t=0}^{T-1}\mathbb E\|\nabla f(\bar W_t)\|_*
\le
\frac{\chi_q C_\tau^{\mathrm{NS}}}{(KT)^{1/3}},
\label{dmx:eq:ns-rate}
\end{equation}
where $\chi_q$ is defined in Lemma~\ref{dmx:lem:ns}, and $C_\tau^{\mathrm{NS}}$, given explicitly in \eqref{dmx:eq:ns-constant} of Appendix~\ref{dmx:app:ns}, is independent of $K,T$ under the stated conditions.
\end{theorem}
Two remarks are worth noting. First, by Lemma~\ref{lem:reuse}, the NS factor $\chi_q$ rapidly approaches $1$ as $q$ increases, so \alg with finite-step NS approaches the exact-orthogonalization rate while avoiding its computational cost. 
Second, compared to FedMuon-SF \cite{takezawa2026fedmuon}, whose analysis only establishes convergence for an arbitrary number of NS iterations under fixed parameters, our convergence result is stronger in two aspects: 1) We provide a quantitative characterization of the discrepancy between the exact orthogonalization and its finite-step NS approximation, rather than only proving convergence in a qualitative sense; and 2) We show that our method still preserves the $\mathcal{O}((KT)^{-1/3})$ convergence rate, which is stronger than the rate established in FedMuon-SF.
Meanwhile, the other two FedMuon works \cite{liu2025fedmuon,zhang2025provable} only analyze the exact orthogonalization version and therefore do not capture the effect of practical inexact orthogonalization.

\section{\alg with Low-Rank Communications}
\label{dmx:sec:lr}\label{sec:fed3}

In this section, we focus on a communication-efficient variant of \algn, which is termed as \alglr and designed for bandwidth-constrained distributed learning.
The goal is to reduce communication cost and quantify the additional error it introduces into the stationarity bound.

\paragraph{1) Algorithm Description.}
In each communication round, the nodes do not transmit full matrix-valued momentum updates.
Instead, each node $k$ computes a rank-$r$ truncated SVD of its local STORM estimator $\widetilde M_{t+1}^{(k)}$, giving
$M_{t+1,r}^{(k)}=U_{t+1,r}^{(k)}\Sigma_{t+1,r}^{(k)}(V_{t+1,r}^{(k)})^\top$, and uploads only the factorized representation $(U_{t+1,r}^{(k)},\Sigma_{t+1,r}^{(k)},V_{t+1,r}^{(k)})$ to the server.
Meanwhile, nodes upload their local models $W_{t+1}^{(k)}$.

Upon receiving the factorized updates, the server first reconstructs the matrices and performs linear aggregation as follows:
$\bar M_{t+1,r}=\frac1K\sum_{k=1}^K M_{t+1,r}^{(k)}=\frac1K\sum_{k=1}^K U_{t+1,r}^{(k)}\Sigma_{t+1,r}^{(k)}(V_{t+1,r}^{(k)})^\top$.
Performing this reconstruction step first is essential, as low-rank factorization is {\em not} a linear operation and breaks commutativity.
The server then applies a rank-$r$ truncated SVD to $\bar M_{t+1,r}$, giving
$M_{t+1,r}=U_{t+1,r}\Sigma_{t+1,r}V_{t+1,r}^\top$, and broadcasts only $(U_{t+1,r},\Sigma_{t+1,r},V_{t+1,r})$ to all nodes.
Also, the server computes the averaged model
$\bar W_{t+1} = \frac{1}{K}\sum_k W_{t+1}^{(k)}$
and broadcasts it to the nodes.
Upon receiving the information from the server, each node reconstructs the synchronized momentum as
$M_{t+1}^{(k)}\leftarrow M_{t+1,r}=U_{t+1,r}\Sigma_{t+1,r}V_{t+1,r}^\top$,
and continues local updates until the next communication round.

\paragraph{2) Convergence Analysis under Low-Rank Communications.}
Low-rank communication introduces truncation errors at both the clients and the server. Let $M_{t,r}^{(k)}$ denote the rank-$r$ approximation of the local momentum estimator $\widetilde M_t^{(k)}$, let $\bar M_{t,r}:=\frac{1}{K}\sum_{k=1}^K M_{t,r}^{(k)}$ denote their average, and let $M_{t,r}$ denote its rank-$r$ recompression at the server.
Since both stages use truncated SVD, the residual is
$
\rho_t:= \frac{1}{K}\sum_{k=1}^{K}
(\sum_{j=r+1}^{d}
\sigma_j(\widetilde M_t^{(k)})^2)^{1/2}+
(\sum_{j=r+1}^{d}
\sigma_j(\bar M_{t,r})^2)^{1/2},
$
where $\sigma_j(A)$ is the $j$th largest singular
value of $A$. Set $\rho_t=0$ outside communication
boundaries.
To account for the propagation of these errors through the recursive momentum estimator, define $R_T:=\frac{1}{(1-\beta)T}\sum_{t=1}^{T-1}(1-\beta^{T-t})\mathbb{E}[\rho_t]$.
Let $\rho_{r,T}:=\max_{1\le t<T}\mathbb{E}[\rho_t]$, with $\rho_{r,T}=0$ when $T=1$.
We now state the convergence result for \algnlr{}.
\begin{theorem}[Effect of Low-Rank Communication]
\label{dmx:thm:lr}
Under the assumptions and parameter choices of
Theorem~\ref{dmf:thm:main} with rank $r$, \algnlr{} with exact
orthogonalization satisfies
$
\frac{1}{T}\sum_{t=0}^{T-1}
\mathbb{E}\|\nabla f(\bar W_t)\|_F
\le
\frac{C_\tau}{(KT)^{1/3}}
+\frac{4\sqrt d\,T^{2/3}}{\tau K^{1/3}}\rho_{r,T},
$
where $C_\tau$ is the constant in
Theorem~\ref{dmf:thm:main}.
\end{theorem}
Low-rank communication adds an error governed by the discarded client and server spectral tails.
The bound retains the $O((KT)^{-1/3})$ rate if $\rho_{r,T}=O(\tau/(\sqrt d\,T))$.
Note that, if $r=d$, the tails vanish, recovering the full-communication guarantee.
The complete proof is provided in Appendix~\ref{dmx:app:lr}.
Unlike FedMuon-LRSF~\cite{liu2025fedmuon}, whose analysis is based on the full-rank update, our bound directly accounts for the rank-$r$ communication error through $R_T$.

The same low-rank communication protocol can be combined with the finite-step NS directions of Section~\ref{dmx:sec:ns}, leading to the following result.
\begin{corollary}[Combined LR--NS Error]
\label{dmx:cor:hybrid}
Under the conditions of Theorems~\ref{dmx:thm:ns} and~\ref{dmx:thm:lr}, with the NS spectral condition imposed on the LR--NS trajectory, we have
$\frac{1}{T}\sum_{t=0}^{T-1}
\mathbb{E}\|\nabla f(\bar W_t)\|_*
\le \chi_q\left(
\frac{C_\tau^{\mathrm{NS}}}{(KT)^{1/3}}
+\frac{4\sqrt d\,T^{2/3}}{\tau K^{1/3}}\rho_{r,T}
\right)$.
\end{corollary}
\begin{remark}[Communication Reduction]
For $M\in\mathbb{R}^{m\times n}$, dense momentum transmission requires $mn$ scalars, whereas rank-$r$ transmission of $(U_r,\Sigma_r,V_r)$ requires $r(m+n+1)$ scalars, since only the diagonal entries of $\Sigma_r$ are transmitted. Thus, when $r\ll \min\{m,n\}$, \algnlr{} substantially reduces the momentum communication cost.
\end{remark}

\section{Experiments}
\label{sec:experiments}

We evaluate \algn{} on image classification tasks to examine its empirical
performance and the practical effects of finite-step Newton--Schulz (NS)
orthogonalization and low-rank communication. Additional implementation
details, hyperparameter settings, and ablation studies are provided in
Appendix~\ref{app:exp_details}.

\subsection{CIFAR-10 Classification with ResNet-20}
\label{sec:exp_cifar}

\textbf{1) Dataset and Setting.}
We evaluate \alg on CIFAR-10~\cite{krizhevsky2009learning} using a widened ResNet-20 with Group Normalization.
We consider $K=5$ fully participating nodes with an IID data partition and report results over three random seeds.
Each node performs \(\tau=5\) local updates per communication round with a batch size of 128.
Detailed training settings are provided in Appendix~\ref{app:cifar_details}.

\textbf{2) Baselines.}
We compare \algn{} with FedAvg~\cite{mcmahan2017communication}, FedAvg-SGDM (FedAvg with momentum), FedMuon-FA~\cite{zhang2025provable} and FedMuon-SF~\cite{takezawa2026fedmuon}.
We also compare four \algn{} variants:
\algn{}-Full SVD with exact orthogonalization and full-rank communication,
\algnns{} with finite-step NS orthogonalization,
\algn{}-Low Rank SVD with low-rank momentum communication,
and \algn{}-Hybrid, which combines low-rank communication with
NS orthogonalization.

\textbf{3) Results.}
Our results highlight three aspects of \algn{}.
First, \algnns{} improves test accuracy from
$86.95\%$ for FedMuon-FA to
$89.92\%$, supporting the benefit of the
STORM-based gradient estimator under the same
epoch budget (Table~\ref{tab:test_accuracy}).
Second, NS achieves comparable accuracy to SVD
($89.92\%$ versus $89.69\%$) while reducing mean training time from 98.61 to 48.88 minutes, a $2.02\times$ speedup (Figure~\ref{fig:alg} and \ref{fig:runtime}).
Third, rank-$32$ compression reduces the
simulated communication payload by $40.5\%$,
from 385.39 to 229.34\,GB.
\algnlr{} and \algn{}-Hybrid achieve
test accuracies of $85.08\%$ and $85.03\%$, respectively,
illustrating the accuracy--communication trade-off (Figure~\ref{fig:r}).
Finally, the median relative direction error (defined in Appendix~\ref{app:cifar_details}) drops
substantially from $q=3$ to $q=5$ and changes little
with further iterations, supporting $q=5$ as a
practical choice for the evaluated matrices (Figure~\ref{fig:q}).

\begin{figure}[t]
\centering
\subfigure[Validation accuracy.]{%
  \includegraphics[width=0.47\linewidth]{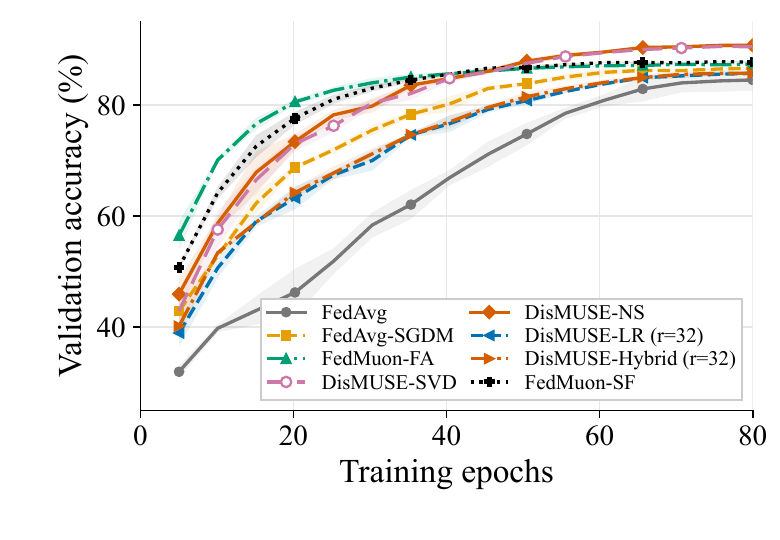}%
  \label{fig:alg}}
\hfill
\subfigure[Operator runtime (ms).]{%
  \includegraphics[width=0.47\linewidth]{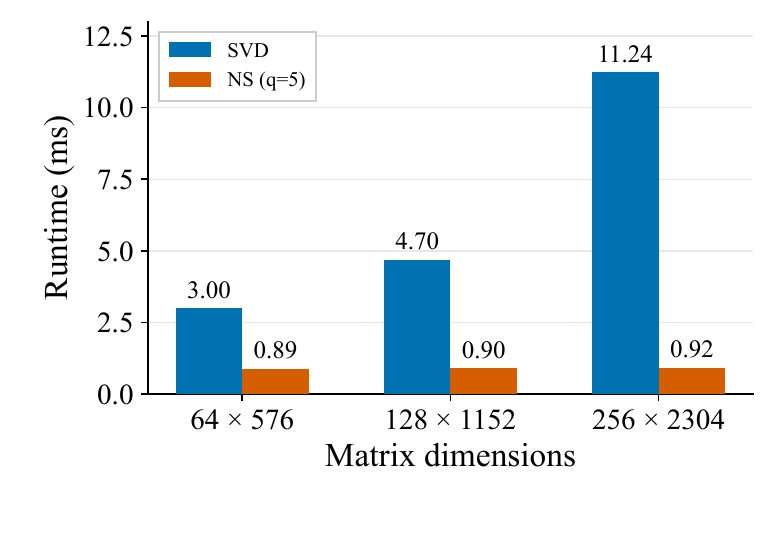}%
  \label{fig:runtime}}
\par\medskip
\subfigure[Low Rank SVD trade-off.]{%
  \includegraphics[width=0.47\linewidth]{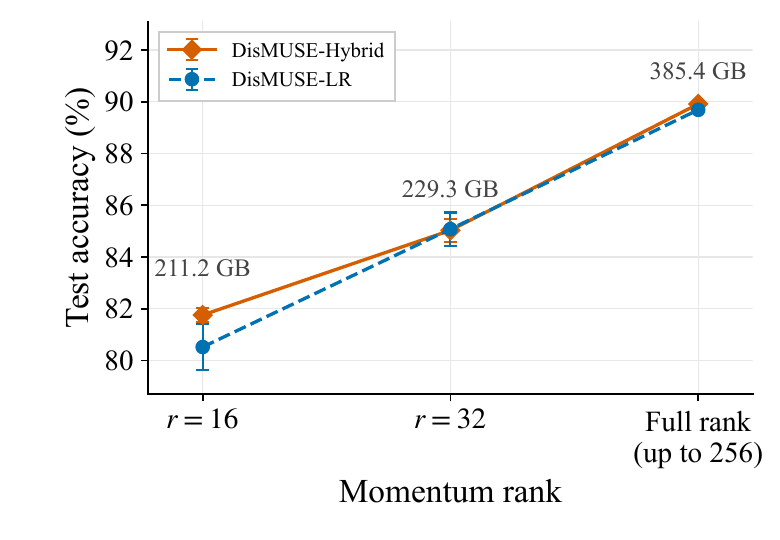}%
  \label{fig:r}}
\hfill
\subfigure[NS relative direction err.]{%
  \includegraphics[width=0.47\linewidth]{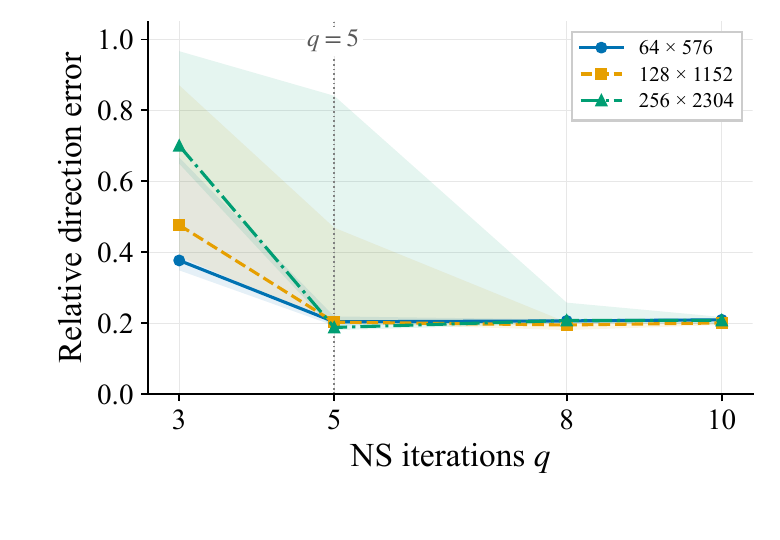}%
  \label{fig:q}}
\caption{CIFAR-10 Classification with ResNet-20}
\label{fig:5agent}\label{fig:exp}
\end{figure}

\subsection{Tiny Shakespeare Language Modeling with a Four-Layer Transformer}
\label{sec:exp_fashion}
\begin{figure}[t]
\centering
\subfigure[Validation loss.]{%
  \includegraphics[width=0.47\linewidth]{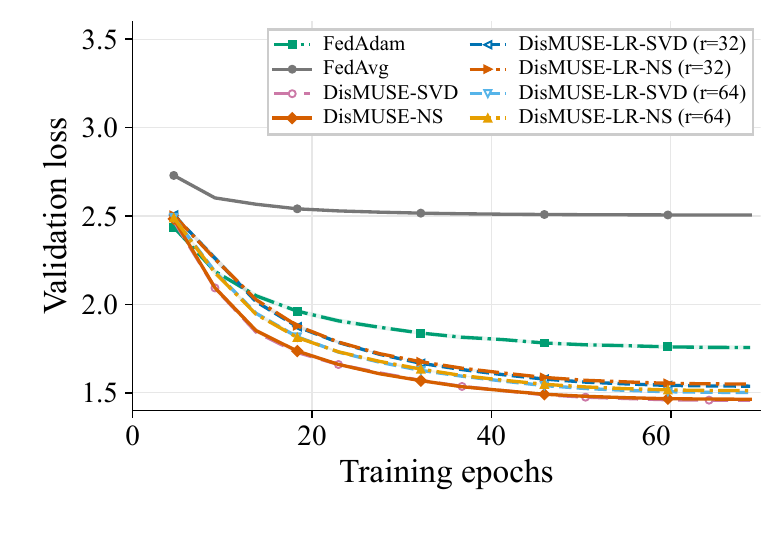}%
  \label{fig:svd}}
\hfill
\subfigure[Operator runtime (ms).]{%
  \includegraphics[width=0.47\linewidth]{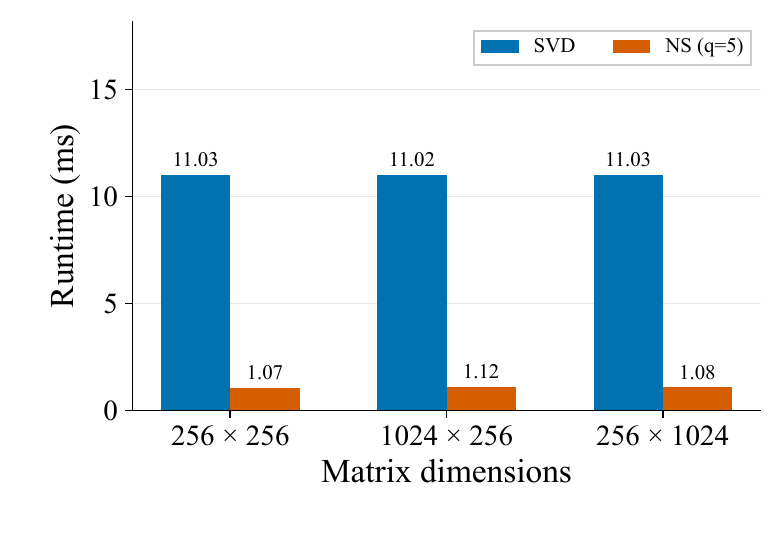}%
  \label{fig:shakespeare}}
\caption{Tiny Shakespeare Language Modeling with a Four-Layer Transformer.}
\label{fig:comp}
\end{figure}
\textbf{1) Dataset and Setting.}
We evaluate \alg on the Tiny Shakespeare dataset~\cite{karpathy2015tinyshakespeare} using a four-layer
causal Transformer for character-level language modeling.
We consider $K=5$ fully participating nodes that independently
sample from a shared training pool and synchronize every five
local updates.
Detailed model architecture and training settings are provided in Appendix~\ref{app:fashion_details}.

\textbf{2) Baselines.}
We compare three variants of \algn{} (exact SVD orthogonalization with full-rank communication, finite-step NS orthogonalization, and low-rank communication) with the FedAvg and FedAdam (FedAvg with local Adam) baselines.

\textbf{3) Results.}
As shown in Figure~\ref{fig:svd}, the NS variant achieves validation loss close to the exact-SVD variant, while the low-rank variant retains competitive performance under compressed communication.
Figure~\ref{fig:shakespeare} shows that NS substantially
reduces the spectral direction computation time compared with exact-SVD.
These results show that the practical NS and low-rank variants retain much of the empirical performance of exact full-rank \algn{}.

\section{Conclusion}
We proposed \algn, a distributed spectral optimization
framework combining Muon-style orthogonalized momentum
with recursive variance reduction. Under the stated
assumptions, \alg achieves an $O((KT)^{-1/3})$
stationarity rate, which finite-step NS orthogonalization
preserves under a uniform normalized support-spectrum
condition. We also derived explicit error bounds for
low-rank communication and its combination with NS,
accounting for accumulated client and server
truncation errors.

\section*{AI Use Statement}
In this work, we did not use generative AI tools for any tasks requiring disclosure.
Additionally, we used generative AI tools to create or modify scientific figures or images, source or search for information, edit a research
paper to improve readability, and format references.
We have reviewed all AI-assisted work.
Literature claims and citations were checked against the original sources,
and all AI-assisted text and figures were reviewed by the authors.
We take responsibility for the final content of this work, including text, claims or artifacts produced with the aid of generative AI.

\bibliographystyle{plainnat}
\bibliography{references}
\clearpage
\appendix
\begin{table}[H]
\centering
\caption{Main parameters used in \algn{}.}
\label{tab:main-parameters}
\renewcommand{\arraystretch}{1.15}
\begin{tabularx}{\linewidth}{c|X}
\toprule
\textbf{Parameter} & \textbf{Definition} \\
\midrule
$K$ & Number of distributed nodes. \\

$T$ & Number of local optimization iterations per node. \\

$\eta$ & Constant model-update stepsize. \\

$\beta$ & STORM momentum parameter, where $\beta\in(0,1)$. \\

$\tau$ & Communication period; synchronization occurs every $\tau$ local steps. \\

$B_0$ & Initial mini-batch size used to initialize the momentum estimator. \\

$d$ & Maximum possible matrix rank, $d=\min\{m,n\}$, in the exact analysis. \\

$r$ & Communication truncation rank in \algnlr{}; local directions need not have rank $r$. \\

$q$ & Number of Newton--Schulz iterations. \\

$L$ & Mean-square stochastic and Frobenius objective smoothness constant. \\

$L_*$ & Spectral/nuclear norm-pair smoothness constant. \\

$\sigma^2$ & Upper bound on the stochastic-gradient variance. \\

$\delta^2$ & Upper bound on gradient heterogeneity across nodes. \\

$\varepsilon_q$ & Uniform error bound between the $q$-step NS direction and exact support polar. \\

$\chi_q$ & NS approximation factor, defined as
$\chi_q=(1-\varepsilon_q)^{-1}$. \\

$\rho_s$ & Total client-side truncation and server-side recompression error at boundary $s$. \\

$R_T$ & Geometrically weighted accumulated compression budget. \\
\bottomrule
\end{tabularx}
\end{table}
\paragraph{Proof roadmap.}
For exact orthogonalization, Lemma~\ref{dmf:lem:descent} establishes the one-step descent inequality. Lemmas~\ref{dmf:lem:model}, \ref{dmf:lem:momentum}, and \ref{dmf:lem:estimation} bound the model-consensus error, momentum-consensus error, and averaged STORM estimation error, respectively. Substituting these bounds into the descent inequality and telescoping over $t$ gives Theorem~\ref{dmf:thm:main}.

For finite-step Newton--Schulz orthogonalization, Lemma~\ref{dmx:lem:ns} relates the number of NS steps $q$ to the approximation error $\varepsilon_q$ and the factor $\chi_q$ under the uniform normalized support-spectrum condition. Lemmas~\ref{dmx:lem:ns-descent} and~\ref{dmx:lem:ns-errors} establish the descent and error bounds, yielding Theorem~\ref{dmx:thm:ns}.

For low-rank communication, the proof separates the client and server truncation residuals from the centered stochastic estimation error. Their accumulated budget $R_T$ gives Theorem~\ref{dmx:thm:lr} and the combined LR--NS error bound in Corollary~\ref{dmx:cor:hybrid}.

\section{Proof of Theorem~\ref{dmf:thm:main}}
\label{dmf:app:main}

\paragraph{Smoothness constant.}
We use the same constant $L$ for stochastic and objective smoothness. By Assumptions~\ref{ass: smooth} and~\ref{dmf:ass:variance},
\[
\begin{aligned}
\|\nabla f^{(k)}(W)-\nabla f^{(k)}(W')\|_F^2
&=\|\mathbb E_\xi[\nabla f^{(k)}(W;\xi)-\nabla f^{(k)}(W';\xi)]\|_F^2
\\
&\le\mathbb E_\xi\|\nabla f^{(k)}(W;\xi)-\nabla f^{(k)}(W';\xi)\|_F^2
\le L^2\|W-W'\|_F^2.
\end{aligned}
\]
For a fresh sample independent of the preceding history, the same mean-square bound applies conditionally at the two history-measurable model positions used in the STORM update.

\paragraph{Algorithm definitions used in the proof.}
Let $d=\min\{m,n\}$. The positive-singular-value polar factor, with $\Orth(0)=0$, satisfies
\begin{equation}\label{dmf:eq:polar}
\|\Orth(M)\|_{\mathrm{op}}\le1,\qquad
\|\Orth(M)\|_F\le\sqrt d,\qquad
\langle M,\Orth(M)\rangle=\|M\|_*.
\end{equation}
The dense initialization in Algorithm~\ref{alg:fedmuon_unified} is
\begin{equation}\label{dmf:eq:init}
M_0^{(k)}=\bar M_0=\frac1K\sum_{j=1}^K\frac1{B_0}\sum_{i=1}^{B_0}\nabla f^{(j)}(W_0;\xi_{0,i}^{(j)}).
\end{equation}
For the proof, let $Y_{t+1}^{(k)}$ denote the local model before any averaging in Algorithm~\ref{alg:fedmuon_unified}; after averaging, it is denoted by $W_{t+1}^{(k)}$. The spectral direction and candidate model update are
\begin{equation}\label{dmf:eq:local-model}
O_t^{(k)}=\begin{cases}
\Orth(M_t^{(k)}), & \text{exact orthogonalization},\\
X_{t,q}^{(k)}, & \text{$q$-step NS as defined in Section~\ref{dmx:sec:ns}},
\end{cases}
\qquad Y_{t+1}^{(k)}=W_t^{(k)}-\eta O_t^{(k)}.
\end{equation}

For integers $K,T,B_0\ge1$, $\tau\in\{1,\ldots,T\}$, $\eta>0$, and $\beta\in(0,1)$, the general bound underlying Theorem~\ref{dmf:thm:main} uses the variance budget
\begin{equation}
\mathcal V_T:=\frac{\sigma^2}{KB_0(1-\beta)T}
 +\frac{2(1-\beta)\sigma^2}{K} +\frac{2L^2\eta^2d(4\tau-3)}{K(1-\beta)}.
\label{dmf:eq:variance-budget}
\end{equation}
Let $\Delta:=f(W_0)-f_*$. Under the assumptions and updates of that theorem,
\begin{equation}
\begin{aligned}
\frac1T\sum_{t=0}^{T-1}\mathbb E\|\nabla f(\bar W_t)\|_F
\le{}&\frac{\Delta}{\eta T}+\frac{\eta Ld}{2}+4\eta Ld(\tau-1)+4\beta\eta Ld(\tau-1)\\
&+8(1-\beta)\eta Ld(\tau-1)^2+2\sqrt d(1-\beta)(\tau-1)(\delta+2\sigma)+2\sqrt{d\mathcal V_T}.
\end{aligned}
\label{dmf:eq:main-bound}
\end{equation}

Throughout this section, let $a=1-\beta$, $h=\tau-1$, and
\[
H_t:=\frac1K\sum_{k=1}^K\nabla f^{(k)}(W_t^{(k)}),\qquad
\bar M_t:=\frac1K\sum_{k=1}^K M_t^{(k)}.
\]
All variables at a synchronization boundary refer to the states after
the corresponding averaging operations. We first establish a descent
inequality, followed by bounds on the model-consensus error,
momentum-consensus error, and averaged estimation error.

\begin{lemma}[One-step descent]
\label{dmf:lem:descent}
Suppose Assumption~\ref{dmf:ass:smooth} holds. The averaged model satisfies
\begin{align}
f(\bar W_{t+1})
\le{}&f(\bar W_t)-\eta\|\nabla f(\bar W_t)\|_F
 +\frac{2\eta\sqrt d\,L}{K}\sum_{k=1}^K
     \|\bar W_t-W_t^{(k)}\|_F\nonumber\\
&+2\eta\sqrt d\,\|H_t-\bar M_t\|_F
 +\frac{2\eta\sqrt d}{K}\sum_{k=1}^K
     \|\bar M_t-M_t^{(k)}\|_F
 +\frac{\eta^2Ld}{2}.
\label{dmf:eq:descent}
\end{align}
\end{lemma}

\begin{proof}
By the update rule and the fact that averaging preserves the mean model,
\[
\bar W_{t+1}-\bar W_t
=-\eta\frac1K\sum_{k=1}^K O_t^{(k)}.
\]
The $L$-smoothness of $f$ implies
\begin{align}
f(\bar W_{t+1})
\le{}&f(\bar W_t)
 -\frac{\eta}{K}\sum_{k=1}^K
   \langle\nabla f(\bar W_t),O_t^{(k)}\rangle
 +\frac{\eta^2L}{2}
   \left\|\frac1K\sum_{k=1}^K O_t^{(k)}\right\|_F^2.
\label{dmf:eq:descent-smooth}
\end{align}
For each node $k$, the spectral--nuclear norm duality and
\eqref{dmf:eq:polar} give
\begin{align*}
\langle\nabla f(\bar W_t),O_t^{(k)}\rangle
&=\langle M_t^{(k)},O_t^{(k)}\rangle
 +\langle\nabla f(\bar W_t)-M_t^{(k)},O_t^{(k)}\rangle\\
&\ge\|M_t^{(k)}\|_*
 -\|\nabla f(\bar W_t)-M_t^{(k)}\|_*\\
&\ge\|\nabla f(\bar W_t)\|_*
 -2\|\nabla f(\bar W_t)-M_t^{(k)}\|_*\\
&\ge\|\nabla f(\bar W_t)\|_F
 -2\sqrt d\,\|\nabla f(\bar W_t)-M_t^{(k)}\|_F.
\end{align*}
Moreover, Jensen's inequality yields
\[
\left\|\frac1K\sum_{k=1}^K O_t^{(k)}\right\|_F^2
\le\frac1K\sum_{k=1}^K\|O_t^{(k)}\|_F^2\le d.
\]
To bound the gradient--momentum discrepancy, we insert $H_t$ and $\bar M_t$:
\begin{align*}
\|\nabla f(\bar W_t)-M_t^{(k)}\|_F
\le{}&\|\nabla f(\bar W_t)-H_t\|_F
 +\|H_t-\bar M_t\|_F+\|\bar M_t-M_t^{(k)}\|_F.
\end{align*}
By Assumption~\ref{dmf:ass:smooth},
\[
\|\nabla f(\bar W_t)-H_t\|_F
\le\frac1K\sum_{k=1}^K
\|\nabla f^{(k)}(\bar W_t)-\nabla f^{(k)}(W_t^{(k)})\|_F
\le\frac L K\sum_{k=1}^K\|\bar W_t-W_t^{(k)}\|_F.
\]
Substituting these bounds into \eqref{dmf:eq:descent-smooth} gives
\eqref{dmf:eq:descent}, which completes the proof.
\end{proof}

\begin{lemma}[Model-consensus error and model displacement]
\label{dmf:lem:model}
For the specified updates and synchronization rules, the following bounds
hold pathwise:
\begin{equation}
\frac1K\sum_{k=1}^K\|\bar W_t-W_t^{(k)}\|_F
\le2\eta(\tau-1)\sqrt d,
\label{dmf:eq:model-consensus}
\end{equation}
and
\begin{equation}
\frac1T\sum_{t=0}^{T-1}\frac1K\sum_{k=1}^K
\|W_{t+1}^{(k)}-W_t^{(k)}\|_F^2
\le\eta^2d(4\tau-3).
\label{dmf:eq:model-displacement}
\end{equation}
\end{lemma}

\begin{proof}
Let $s_t=\lfloor t/\tau\rfloor$ and $\ell_t=t-s_t\tau\in\{0,\ldots,\tau-1\}$.
At the beginning of the current communication period,
$W_{s_t\tau}^{(k)}=\bar W_{s_t\tau}$ for all $k$.
Unrolling the local updates within this period gives
\begin{align*}
W_t^{(k)}&=\bar W_{s_t\tau}
 -\eta\sum_{j=s_t\tau}^{t-1}O_j^{(k)},\\
\bar W_t&=\bar W_{s_t\tau}
 -\eta\sum_{j=s_t\tau}^{t-1}\frac1K\sum_{k'=1}^K O_j^{(k')}.
\end{align*}
Consequently, for each node $k$,
\begin{align}
\|W_t^{(k)}-\bar W_t\|_F
&\le\eta\sum_{j=s_t\tau}^{t-1}
 \left(\|O_j^{(k)}\|_F+
 \left\|\frac1K\sum_{k'=1}^K O_j^{(k')}\right\|_F\right)
\le2\eta\ell_t\sqrt d.
\label{dmf:eq:individual-drift}
\end{align}
Averaging over $k$ proves \eqref{dmf:eq:model-consensus}.

Next, we distinguish local steps from synchronization steps. If
$(t+1)\bmod\tau\ne0$, then
\[
\|W_{t+1}^{(k)}-W_t^{(k)}\|_F
=\eta\|O_t^{(k)}\|_F\le\eta\sqrt d.
\]
If $(t+1)\bmod\tau=0$, then $\ell_t=\tau-1$ and
\[
W_{t+1}^{(k)}-W_t^{(k)}
=\bar W_t-W_t^{(k)}-\eta\frac1K\sum_{k'=1}^K O_t^{(k')}.
\]
Using \eqref{dmf:eq:individual-drift}, we obtain
\[
\|W_{t+1}^{(k)}-W_t^{(k)}\|_F
\le(2\tau-1)\eta\sqrt d.
\]
There are $N_T=\lfloor T/\tau\rfloor$ synchronization steps among
the $T$ transitions. Hence
\begin{align*}
\frac1T\sum_{t=0}^{T-1}\frac1K\sum_{k=1}^K
\|W_{t+1}^{(k)}-W_t^{(k)}\|_F^2
&\le\eta^2d\left(1+\frac{N_T}{T}\big((2\tau-1)^2-1\big)\right)\\
&\le\eta^2d\left(1+\frac{(2\tau-1)^2-1}{\tau}\right)
=\eta^2d(4\tau-3).
\end{align*}
The last inequality also holds when the final communication period is
incomplete, since $N_T/T\le1/\tau$ and $(2\tau-1)^2-1\ge0$.
\end{proof}

\begin{lemma}[Consensus error of the STORM estimator]
\label{dmf:lem:momentum}
Suppose Assumptions~\ref{dmf:ass:smooth}--\ref{dmf:ass:heterogeneity}
hold. For every $t$, the momentum-consensus error satisfies
\begin{align}
\frac1K\sum_{k=1}^K\mathbb E\|\bar M_t-M_t^{(k)}\|_F
\le (1-\beta)(\tau-1)(\delta+2\sigma) +4(1-\beta)L\eta(\tau-1)^2\sqrt d
 +2\beta L\eta(\tau-1)\sqrt d.
\label{dmf:eq:momentum-consensus}
\end{align}
\end{lemma}

\begin{proof}
Define
\[
c_t:=\frac1K\sum_{k=1}^K\mathbb E\|\bar W_t-W_t^{(k)}\|_F,
\qquad
d_t:=\frac1K\sum_{k=1}^K\mathbb E\|\bar M_t-M_t^{(k)}\|_F.
\]
The synchronized initialization and the averaging rule give $d_{s\tau}=0$
for every communication boundary. It therefore suffices to bound the
increments within a communication period.

Consider an iteration with $(t+1)\bmod\tau\ne0$, and write
\begin{align*}
G_{t+1}^{(k)}&:=\nabla f^{(k)}(W_{t+1}^{(k)};\xi_{t+1}^{(k)}),\\
Q_{t+1}^{(k)}&:=\nabla f^{(k)}(W_{t+1}^{(k)};\xi_{t+1}^{(k)})
 -\nabla f^{(k)}(W_t^{(k)};\xi_{t+1}^{(k)}).
\end{align*}
Let $\bar G_{t+1}$ and $\bar Q_{t+1}$ denote their node averages.
The local estimator update becomes
$M_{t+1}^{(k)}=\beta M_t^{(k)}+aG_{t+1}^{(k)}+\beta Q_{t+1}^{(k)}$.
Subtracting its average and taking expected Frobenius norms yields
\begin{equation}
d_{t+1}\le\beta d_t
 +\frac aK\sum_{k=1}^K\mathbb E\|G_{t+1}^{(k)}-\bar G_{t+1}\|_F
 +\frac\beta K\sum_{k=1}^K\mathbb E\|Q_{t+1}^{(k)}-\bar Q_{t+1}\|_F.
\label{dmf:eq:consensus-recursion-raw}
\end{equation}

For the first additional term, separate the stochastic noise from the
true gradients. By Assumption~\ref{dmf:ass:variance} and Jensen's inequality,
the average expected norm of the centered noise is at most $2\sigma$.
For the true-gradient differences, insert the gradients at
$\bar W_{t+1}$ and use Assumptions~\ref{dmf:ass:smooth} and
\ref{dmf:ass:heterogeneity} to obtain
\begin{align*}
&\frac1K\sum_{k=1}^K
\mathbb E\|\nabla f^{(k)}(W_{t+1}^{(k)})-H_{t+1}\|_F\\
&\quad\le
\frac1K\sum_{k=1}^K\mathbb E
\|\nabla f^{(k)}(W_{t+1}^{(k)})-\nabla f^{(k)}(\bar W_{t+1})\|_F\\
&\qquad+
\frac1K\sum_{k=1}^K\mathbb E
\|\nabla f^{(k)}(\bar W_{t+1})-\nabla f(\bar W_{t+1})\|_F
 +\mathbb E\|\nabla f(\bar W_{t+1})-H_{t+1}\|_F\\
&\quad\le2Lc_{t+1}+\delta.
\end{align*}
Thus,
\begin{equation}
\frac1K\sum_{k=1}^K\mathbb E\|G_{t+1}^{(k)}-\bar G_{t+1}\|_F
\le2Lc_{t+1}+\delta+2\sigma.
\label{dmf:eq:current-gradient-disagreement}
\end{equation}
The heterogeneity bound is used after averaging over nodes; no
individual-node bound by $\delta$ is required.

For the second additional term, conditional mean-square smoothness and
Jensen's inequality give
\[
\mathbb E\|Q_{t+1}^{(k)}\|_F
\le L\mathbb E\|W_{t+1}^{(k)}-W_t^{(k)}\|_F
\le L\eta\sqrt d.
\]
The last inequality holds because the current transition is not a
synchronization step. Therefore,
\begin{equation}
\frac1K\sum_{k=1}^K\mathbb E\|Q_{t+1}^{(k)}-\bar Q_{t+1}\|_F
\le2L\eta\sqrt d.
\label{dmf:eq:difference-disagreement}
\end{equation}
Combining \eqref{dmf:eq:consensus-recursion-raw}--\eqref{dmf:eq:difference-disagreement}
and Lemma~\ref{dmf:lem:model} gives
\[
d_{t+1}\le\beta d_t
 +a(4L\eta h\sqrt d+\delta+2\sigma)+2\beta L\eta\sqrt d.
\]
There are at most $h$ local transitions after the most recent synchronized
state. Unrolling the recursion from $d_{s\tau}=0$ and using
$\sum_{j=0}^{h-1}\beta^j\le h$ yields
\[
d_t\le h\bigl[a(4L\eta h\sqrt d+\delta+2\sigma)
 +2\beta L\eta\sqrt d\bigr].
\]
At a synchronization boundary $d_t=0$, so the same bound holds there.
Substituting $a=1-\beta$ and $h=\tau-1$ completes the proof.
\end{proof}

\begin{lemma}[Averaged STORM estimation error]
\label{dmf:lem:estimation}
Suppose Assumptions~\ref{dmf:ass:smooth}--\ref{dmf:ass:heterogeneity} hold. Under the specified initialization
and synchronization rules, for any $T\ge1$,
\begin{equation}
\frac1T\sum_{t=0}^{T-1}\mathbb E\|H_t-\bar M_t\|_F
\le\sqrt{
\frac{\sigma^2}{KB_0(1-\beta)T}
 +\frac{2(1-\beta)\sigma^2}{K}
 +\frac{2L^2\eta^2d(4\tau-3)}{K(1-\beta)}}.
\label{dmf:eq:estimation-bound}
\end{equation}
\end{lemma}

\begin{proof}
Let $\mathcal F_t$ denote the history before drawing
$\{\xi_{t+1}^{(k)}\}_{k=1}^K$. Both $W_t^{(k)}$ and $W_{t+1}^{(k)}$
are $\mathcal F_t$-measurable: the latter is obtained from the existing
models and momentum estimators before any new samples are drawn.
Define
\[
e_t:=\bar M_t-H_t,\qquad
\bar a_t:=\mathbb E\|e_t\|_F^2.
\]
Since averaging the candidate estimators preserves their mean,
\eqref{dmf:eq:storm} implies, including at synchronization boundaries,
\begin{align}
e_{t+1}
={}&\beta e_t+\frac1K\sum_{k=1}^K
\bigl(a\zeta_{t+1}^{(k)}+\beta\upsilon_{t+1}^{(k)}\bigr),
\label{dmf:eq:error-identity}
\end{align}
where
\begin{align*}
\zeta_{t+1}^{(k)}
&:=\nabla f^{(k)}(W_{t+1}^{(k)};\xi_{t+1}^{(k)})
 -\nabla f^{(k)}(W_{t+1}^{(k)}),\\
\upsilon_{t+1}^{(k)}
&:=\nabla f^{(k)}(W_{t+1}^{(k)};\xi_{t+1}^{(k)})
 -\nabla f^{(k)}(W_t^{(k)};\xi_{t+1}^{(k)})\\
&\hspace{1.2em}
 -\bigl(\nabla f^{(k)}(W_{t+1}^{(k)})-\nabla f^{(k)}(W_t^{(k)})\bigr).
\end{align*}
By conditional unbiasedness, both $\zeta_{t+1}^{(k)}$ and
$\upsilon_{t+1}^{(k)}$ have zero conditional mean given $\mathcal F_t$.
The innovations from different nodes are conditionally independent.
Therefore, the cross term with $e_t$ and all cross-node terms vanish,
and taking conditional second moments in \eqref{dmf:eq:error-identity} gives
\begin{align}
\mathbb E[\|e_{t+1}\|_F^2\mid\mathcal F_t]
={}&\beta^2\|e_t\|_F^2
 +\frac1{K^2}\sum_{k=1}^K
\mathbb E[\|a\zeta_{t+1}^{(k)}+\beta\upsilon_{t+1}^{(k)}\|_F^2
 \mid\mathcal F_t]\nonumber\\
\le{}&\beta^2\|e_t\|_F^2
 +\frac{2a^2\sigma^2}{K}
 +\frac{2\beta^2L^2}{K^2}\sum_{k=1}^K
   \|W_{t+1}^{(k)}-W_t^{(k)}\|_F^2.
\label{dmf:eq:conditional-error-recursion}
\end{align}
Here we used $\|A+B\|_F^2\le2\|A\|_F^2+2\|B\|_F^2$ within each
node. We did not assume that $\zeta_{t+1}^{(k)}$ and
$\upsilon_{t+1}^{(k)}$ are independent. For the second term, centering
does not increase the second moment, so
\begin{align*}
\mathbb E[\|\upsilon_{t+1}^{(k)}\|_F^2\mid\mathcal F_t]
&\le\mathbb E\bigl[
\|\nabla f^{(k)}(W_{t+1}^{(k)};\xi_{t+1}^{(k)})
 -\nabla f^{(k)}(W_t^{(k)};\xi_{t+1}^{(k)})\|_F^2
 \mid\mathcal F_t\bigr]\\
&\le L^2\|W_{t+1}^{(k)}-W_t^{(k)}\|_F^2.
\end{align*}
Taking total expectations in \eqref{dmf:eq:conditional-error-recursion}
and using $\beta^2\le1$ in the additive term gives
\begin{equation}
\bar a_{t+1}\le\beta^2\bar a_t+v_t,\qquad
v_t:=\frac{2a^2\sigma^2}{K}
 +\frac{2L^2}{K^2}\sum_{k=1}^K
   \mathbb E\|W_{t+1}^{(k)}-W_t^{(k)}\|_F^2.
\label{dmf:eq:error-recursion}
\end{equation}
The independent initialization in \eqref{dmf:eq:init} yields
$\bar a_0\le\sigma^2/(KB_0)$.
Unrolling \eqref{dmf:eq:error-recursion}, we have
\[
\bar a_t\le\beta^{2t}\bar a_0
 +\sum_{j=0}^{t-1}\beta^{2(t-1-j)}v_j.
\]
Summing over $t=0,\ldots,T-1$, using nonnegativity of $v_j$, and applying
$\sum_{j=0}^{\infty}\beta^{2j}=(1-\beta^2)^{-1}\le a^{-1}$, we obtain
\begin{align*}
\frac1T\sum_{t=0}^{T-1}\bar a_t
&\le\frac{\bar a_0}{aT}+\frac1{aT}\sum_{j=0}^{T-1}v_j\\
&\le\frac{\sigma^2}{KB_0aT}+\frac{2a\sigma^2}{K}
 +\frac{2L^2}{aK^2T}\sum_{j=0}^{T-1}\sum_{k=1}^K
   \mathbb E\|W_{j+1}^{(k)}-W_j^{(k)}\|_F^2\\
&\le\frac{\sigma^2}{KB_0aT}+\frac{2a\sigma^2}{K}
 +\frac{2L^2\eta^2d(4\tau-3)}{aK},
\end{align*}
where the last inequality follows from Lemma~\ref{dmf:lem:model}.
Finally, Jensen's inequality gives
\[
\frac1T\sum_{t=0}^{T-1}\mathbb E\|H_t-\bar M_t\|_F
\le\sqrt{\frac1T\sum_{t=0}^{T-1}\bar a_t}.
\]
Substituting $a=1-\beta$ completes the proof.
\end{proof}

\begin{proof}[Proof of Theorem~\ref{dmf:thm:main}]
Taking expectations in Lemma~\ref{dmf:lem:descent}, we have
\begin{align}
\mathbb E f(\bar W_{t+1})
\le{}&\mathbb E f(\bar W_t)-\eta\mathbb E\|\nabla f(\bar W_t)\|_F
 +\frac{2\eta\sqrt d\,L}{K}\sum_{k=1}^K
   \mathbb E\|\bar W_t-W_t^{(k)}\|_F\nonumber\\
&+2\eta\sqrt d\,\mathbb E\|H_t-\bar M_t\|_F
 +\frac{2\eta\sqrt d}{K}\sum_{k=1}^K
   \mathbb E\|\bar M_t-M_t^{(k)}\|_F
 +\frac{\eta^2Ld}{2}.
\label{dmf:eq:expected-descent}
\end{align}
We bound the model-consensus and momentum-consensus terms separately.
By Lemma~\ref{dmf:lem:model},
\[
\frac{2\eta\sqrt d\,L}{K}\sum_{k=1}^K
\mathbb E\|\bar W_t-W_t^{(k)}\|_F
\le4\eta^2Ldh.
\]
By Lemma~\ref{dmf:lem:momentum},
\begin{align*}
\frac{2\eta\sqrt d}{K}\sum_{k=1}^K
\mathbb E\|\bar M_t-M_t^{(k)}\|_F
&\le2\eta\sqrt d\,a h(\delta+2\sigma)
 +8aL\eta^2dh^2+4\beta L\eta^2dh.
\end{align*}
Substituting these bounds into \eqref{dmf:eq:expected-descent}, summing
over $t=0,\ldots,T-1$, and rearranging gives
\begin{align*}
\eta\sum_{t=0}^{T-1}\mathbb E\|\nabla f(\bar W_t)\|_F
\le{}&f(W_0)-\mathbb E f(\bar W_T)
 +4\eta^2LdhT+\frac{\eta^2LdT}{2}\\
&+2\eta\sqrt d\,a h(\delta+2\sigma)T
 +8aL\eta^2dh^2T+4\beta L\eta^2dhT\\
&+2\eta\sqrt d\sum_{t=0}^{T-1}\mathbb E\|H_t-\bar M_t\|_F.
\end{align*}
Dividing by $\eta T$, using $f(\bar W_T)\ge f_*$, and applying
Lemma~\ref{dmf:lem:estimation}, we obtain
\begin{align*}
\frac1T\sum_{t=0}^{T-1}\mathbb E\|\nabla f(\bar W_t)\|_F
\le{}&\frac{\Delta}{\eta T}+4\eta Ldh+\frac{\eta Ld}{2}
 +2\sqrt d\,a h(\delta+2\sigma)
 +8aL\eta dh^2+4\beta L\eta dh\\
&+2\sqrt d\sqrt{
\frac{\sigma^2}{KB_0aT}+\frac{2a\sigma^2}{K}
 +\frac{2L^2\eta^2d(4\tau-3)}{aK}}.
\end{align*}
This is \eqref{dmf:eq:main-bound}.

It remains to establish \eqref{dmf:eq:main-rate}.
Fix $\tau$ independently of $K,T$, assume $T\ge\max\{K^2,\tau\}$,
and choose the parameters in Theorem~\ref{dmf:thm:main}.
Let $u=K^{1/3}T^{-2/3}$ and $\varepsilon_T=(KT)^{-1/3}$.
Then $\eta=u$, $a=u/2$, and
\[
\frac1{\eta T}=\varepsilon_T,\qquad
\frac uK=\varepsilon_T^2,\qquad
u\le\varepsilon_T\le1.
\]
The last inequality uses $T\ge K^2$.
For the estimation-error terms,
\begin{align*}
\frac{\sigma^2}{KB_0aT}
&\le2\sigma^2K^{-4/3}T^{-2/3}
 \le2\sigma^2\varepsilon_T^2,\\
\frac{2a\sigma^2}{K}
&=\sigma^2\varepsilon_T^2,\\
\frac{2L^2\eta^2d(4\tau-3)}{aK}
&=4L^2d(4\tau-3)\varepsilon_T^2.
\end{align*}
Furthermore, $\eta\le\varepsilon_T$ and
$8aL\eta dh^2=4Ld h^2u^2\le4Ld h^2\varepsilon_T$.
Using $\beta\le1$ in the remaining terms therefore gives
\[
\frac1T\sum_{t=0}^{T-1}\mathbb E\|\nabla f(\bar W_t)\|_F
\le C_\tau\varepsilon_T,
\]
where one valid constant is
\begin{align*}
C_\tau=\Delta
 +d\left(\frac L2+4Lh+4Lh+4Lh^2\right)
 +\sqrt d\,h(\delta+2\sigma) +2\sqrt d\sqrt{3\sigma^2+4L^2d(4\tau-3)},
\qquad h=\tau-1.
\end{align*}
For fixed $\tau$ and uniformly bounded problem parameters, this constant
is independent of $K,T$, which proves the claimed order.

All expectations used above are finite. Indeed, the bounded spectral
directions and model averaging keep the models in a deterministic bounded
ball over every finite horizon. Objective smoothness bounds their true
gradients on this ball, and the bounded-variance assumption gives finite
second moments for the stochastic gradients and recursively for the
momentum estimators. Thus the conditional expectations, finite sums, and
Jensen inequalities in the proof are well defined.
\end{proof}

\section{Proof of Theorem~\ref{dmx:thm:ns}}
\label{dmx:app:ns}

The explicit budget in Theorem~\ref{dmx:thm:ns} is, with $a=1-\beta$, $h=\tau-1$, and $\mathcal V_T$ defined in \eqref{dmf:eq:variance-budget},
\begin{equation}
B_T^{\mathrm{NS}}:=\frac{\Delta}{\eta T}
 +\frac{\eta L_*}{2}+4\eta L_*h +4\beta\eta Ld h +8a\eta Ld h^2 +2\sqrt d\,ah(\delta+2\sigma)+2\sqrt{d\mathcal V_T}.
\label{dmx:eq:ns-budget}
\end{equation}

Under the assumptions and NS updates of Theorem~\ref{dmx:thm:ns}, for integers $K,T,B_0\ge1$, $\tau\in\{1,\ldots,T\}$, $\eta>0$, and $\beta\in(0,1)$, the general finite-horizon bound is
\begin{equation}
\frac1T\sum_{t=0}^{T-1}\mathbb E\|\nabla f(\bar W_t)\|_*
\le\chi_qB_T^{\mathrm{NS}},
\label{dmx:eq:ns-bound}
\end{equation}

Throughout this section, $a=1-\beta$, $h=\tau-1$, and
$H_t=K^{-1}\sum_{k=1}^K\nabla f^{(k)}(W_t^{(k)})$.
Using the NS approximation bounds in Lemma~\ref{dmx:lem:ns}, we first establish the associated descent
inequality, and then bound the consensus and estimation errors.

\begin{lemma}[One-step descent with NS directions]
\label{dmx:lem:ns-descent}
Under Assumptions~\ref{dmf:ass:smooth}--\ref{dmf:ass:heterogeneity}, spectral/nuclear norm-pair smoothness, and the NS spectral condition,
the NS updates satisfy, pathwise,
\begin{equation}
\begin{aligned}
f(\bar W_{t+1})\le{}&f(\bar W_t)
 -\eta(1-\varepsilon_q)\|\nabla f(\bar W_t)\|_*\\
&+\frac{2\eta L_*}{K}\sum_{k=1}^K
 \|\bar W_t-W_t^{(k)}\|_{\mathrm{op}}
 +2\eta\sqrt d\,\|H_t-\bar M_t\|_F\\
&+\frac{2\eta\sqrt d}{K}\sum_{k=1}^K
 \|\bar M_t-M_t^{(k)}\|_F+\frac{\eta^2L_*}{2}.
\end{aligned}
\label{dmx:eq:ns-descent}
\end{equation}
\end{lemma}
\begin{proof}
Write $G_t=\nabla f(\bar W_t)$ and $P_t^{(k)}=\operatorname{Orth}(M_t^{(k)})$.
By Lemma~\ref{dmx:lem:ns} and spectral/nuclear duality,
\[
\langle M_t^{(k)},O_t^{(k)}\rangle
\ge\langle M_t^{(k)},P_t^{(k)}\rangle
 -\|M_t^{(k)}\|_*\|O_t^{(k)}-P_t^{(k)}\|_{\mathrm{op}}
\ge(1-\varepsilon_q)\|M_t^{(k)}\|_*.
\]
Using $\|O_t^{(k)}\|_{\mathrm{op}}\le1$ and $0\le\varepsilon_q<1$, we obtain
\begin{align*}
\langle G_t,O_t^{(k)}\rangle
&\ge(1-\varepsilon_q)\|M_t^{(k)}\|_*
 -\|G_t-M_t^{(k)}\|_*\\
&\ge(1-\varepsilon_q)\|G_t\|_*
 -(2-\varepsilon_q)\|G_t-M_t^{(k)}\|_*\\
&\ge(1-\varepsilon_q)\|G_t\|_*
 -2\|G_t-M_t^{(k)}\|_*.
\end{align*}
Insert $H_t$ and $\bar M_t$ into the last discrepancy. Since
$\|A\|_*\le\sqrt d\|A\|_F$ for every $m\times n$ matrix,
\[
\|G_t-M_t^{(k)}\|_*
\le\frac{L_*}{K}\sum_{j=1}^K\|\bar W_t-W_t^{(j)}\|_{\mathrm{op}}
 +\sqrt d\|H_t-\bar M_t\|_F
 +\sqrt d\|\bar M_t-M_t^{(k)}\|_F.
\]
Model averaging preserves the mean update:
$\bar W_{t+1}-\bar W_t=-\eta K^{-1}\sum_kO_t^{(k)}$.
Spectral/nuclear norm-pair smoothness gives
\[
f(\bar W_{t+1})\le f(\bar W_t)
 -\frac{\eta}{K}\sum_k\langle G_t,O_t^{(k)}\rangle
 +\frac{\eta^2 L_*}{2}\left\|\frac1K\sum_kO_t^{(k)}\right\|_{\mathrm{op}}^2.
\]
The last norm is at most one. Substitution proves
\eqref{dmx:eq:ns-descent}.
\end{proof}

\begin{lemma}[Consensus and estimation errors with NS]
\label{dmx:lem:ns-errors}
Under the assumptions and algorithm of Theorem~\ref{dmx:thm:ns},
\begin{align}
\frac1K\sum_k\|\bar W_t-W_t^{(k)}\|_{\mathrm{op}}&\le2\eta h,
\label{dmx:eq:ns-model}\\
\frac1K\sum_k\mathbb E\|\bar M_t-M_t^{(k)}\|_F
&\le h\bigl[a(\delta+2\sigma)+4aL\eta h\sqrt d
 +2\beta L\eta\sqrt d\bigr],
\label{dmx:eq:ns-momentum}\\
\frac1T\sum_{t=0}^{T-1}\mathbb E\|H_t-\bar M_t\|_F
&\le\sqrt{\mathcal V_T}.
\label{dmx:eq:ns-estimation}
\end{align}
\end{lemma}
\begin{proof}
At $t=s\tau$, both models and estimators are synchronized. Unrolling
model updates from the most recent boundary, each disagreement is a sum
of at most $h$ differences of directions. Each difference has operator
norm at most two by Lemma~\ref{dmx:lem:ns}, proving
\eqref{dmx:eq:ns-model}.

We verify explicitly the conditions needed to reuse the error bounds in
Appendix~\ref{dmf:app:main}. Lemma~\ref{dmx:lem:ns} gives
$\|O_t^{(k)}\|_F\le\sqrt d$. The proof of
Lemma~\ref{dmf:lem:model} uses only this direction bound and model
averaging, and therefore gives
\[
\frac1K\sum_k\|\bar W_t-W_t^{(k)}\|_F\le2\eta h\sqrt d,
\qquad
\frac1{TK}\sum_{t,k}\mathbb E\|W_{t+1}^{(k)}-W_t^{(k)}\|_F^2
\le\eta^2d(4\tau-3).
\]
The estimator recursion and its synchronized initial state are unchanged.
On nonsynchronization steps the model displacement is at most
$\eta\sqrt d$. The conditional sample assumptions and both stochastic
and objective smoothness assumptions are also unchanged. These are
exactly the inputs to Lemma~\ref{dmf:lem:momentum}, whose proof does not
use exact polar alignment. It gives \eqref{dmx:eq:ns-momentum}.

Finally, model averaging occurs before the fresh paired sample and
estimator averaging preserves its mean. Thus the centered innovation
identity in \eqref{dmf:eq:error-identity} still holds, including at
boundaries. Cross-node conditional independence, initialization variance
$\sigma^2/(KB_0)$, and the displacement bound just established discharge
every hypothesis of the proof of Lemma~\ref{dmf:lem:estimation}. Its
recursion and summation give \eqref{dmx:eq:ns-estimation}.
\end{proof}

\begin{proof}[Proof of Theorem~\ref{dmx:thm:ns}]
Taking expectations in \eqref{dmx:eq:ns-descent}, applying
Lemma~\ref{dmx:lem:ns-errors}, and summing over $t=0,\ldots,T-1$ gives
\begin{align*}
\eta(1-\varepsilon_q)\sum_{t=0}^{T-1}\mathbb E\|\nabla f(\bar W_t)\|_*
\le{}&\Delta+4\eta^2L_*hT+\frac{\eta^2L_*T}{2}
 +2\eta\sqrt d\,T\sqrt{\mathcal V_T}\\
&+2\eta\sqrt d\,Th\bigl[a(\delta+2\sigma)
 +4aL\eta h\sqrt d+2\beta L\eta\sqrt d\bigr].
\end{align*}
Division by $\eta T(1-\varepsilon_q)$ proves \eqref{dmx:eq:ns-bound}.
For the parameter choice in Theorem~\ref{dmf:thm:main}, put
$u=K^{1/3}T^{-2/3}$ and $\varepsilon_T=(KT)^{-1/3}$.
As in the proof of Theorem~\ref{dmf:thm:main},
\[
\eta=u,\quad a=u/2,\quad u\le\varepsilon_T\le1,
\quad \frac1{\eta T}=\varepsilon_T,
\quad
\mathcal V_T\le[3\sigma^2+4L^2d(4\tau-3)]\varepsilon_T^2.
\]
Consequently, $B_T^{\mathrm{NS}}\le C_\tau^{\mathrm{NS}}\varepsilon_T$
with the explicit constant
\begin{equation}
C_\tau^{\mathrm{NS}}=\Delta+\frac{L_*}{2}+4L_*h
 +4Ld h+4Ld h^2+\sqrt d\,h(\delta+2\sigma) +2\sqrt d\sqrt{3\sigma^2+4L^2d(4\tau-3)}.
\label{dmx:eq:ns-constant}
\end{equation}
The constant and $\chi_q$ are independent of $K,T$ under the stated
uniformity conditions. This proves \eqref{dmx:eq:ns-rate} without
changing the stationarity norm during the argument.
All expectations are finite: bounded directions confine the iterates
to a deterministic bounded set for each finite horizon, and the
variance assumption gives finite second moments of the gradient samples
and recursively of the estimators.
\end{proof}

\section{Proof of Theorem~\ref{dmx:thm:lr}}
\label{dmx:app:lr}

We use $a=1-\beta$, $h=\tau-1$, and
$H_t=K^{-1}\sum_k\nabla f^{(k)}(W_t^{(k)})$ throughout.
The proof keeps the local directions and communication compression
separate. Exact support-polar directions are computed from the current
state $M_t^{(k)}$; the compression perturbation changes this state only
at a communication boundary.

\begin{lemma}[Two-sided communication residual]
\label{dmx:lem:residual}
For the protocol in Section~\ref{dmx:sec:lr}, define the matrix residual $E_s^{\mathrm{comp}}:=M_{s,r}-\bar{\widetilde M}_s$ at each boundary $s$, where $\bar{\widetilde M}_s=K^{-1}\sum_k\widetilde M_s^{(k)}$, and set $E_s^{\mathrm{comp}}=0$ away from boundaries. Then
\[
\|E_s^{\mathrm{comp}}\|_F\le\rho_s.
\]
For a matrix $A$ with singular values in nonincreasing order, let $A_r$ be its rank-$r$ truncated-SVD approximation. Then
\[
\|A-A_r\|_F^2=\sum_{i=r+1}^d\sigma_i(A)^2.
\]
\end{lemma}
\begin{proof}
A truncated SVD deletes precisely the singular components with indices
larger than $r$; their Frobenius inner products are zero, giving the
second identity. From the definitions,
\[
E_s^{\mathrm{comp}}=M_{s,r}-\bar{\widetilde M}_s
 =\bigl(M_{s,r}-\bar M_{s,r}\bigr)
 +\frac1K\sum_k\bigl(M_{s,r}^{(k)}-\widetilde M_s^{(k)}\bigr).
\]
The triangle inequality proves the first claim. This argument holds
also at singular-value ties when a fixed measurable tie-breaking rule is used.
\end{proof}

\begin{lemma}[Consensus bounds under compressed synchronization]
\label{dmx:lem:lr-consensus}
Under the assumptions and algorithm of Theorem~\ref{dmx:thm:lr},
the model-disagreement and displacement bounds of
Lemma~\ref{dmf:lem:model}, and the momentum-consensus bound of
Lemma~\ref{dmf:lem:momentum}, remain valid.
\end{lemma}
\begin{proof}
The local direction still satisfies
$\|O_t^{(k)}\|_{\mathrm{op}}\le1$ and
$\|O_t^{(k)}\|_F\le\sqrt d$, regardless of the rank of its input.
The models are averaged exactly using the same timing. Therefore the
unrolling and boundary-displacement arguments in
Lemma~\ref{dmf:lem:model} apply without modification.

At each boundary every node receives the same matrix $M_{s,r}$.
Consequently,
$K^{-1}\sum_k\mathbb E\|M_s^{(k)}-\bar M_s\|_F=0$,
even when $M_{s,r}$ differs from the uncompressed candidate mean.
Between boundaries no compression is performed, and the same local
STORM recursion holds. The proof of Lemma~\ref{dmf:lem:momentum}
only unrolls these nonsynchronization transitions from zero disagreement.
It uses the model-disagreement bound, a local displacement bound
$\eta\sqrt d$, and Assumptions~\ref{dmf:ass:smooth}--\ref{dmf:ass:heterogeneity}.
All of these conditions still hold. Thus, with
$D_t=K^{-1}\sum_k\mathbb E\|M_t^{(k)}-\bar M_t\|_F$,
\[
D_t\le h\bigl[a(\delta+2\sigma)+4aL\eta h\sqrt d
 +2\beta L\eta\sqrt d\bigr].
\]
The possibly biased common momentum state affects the averaged
estimation error, which is treated next.
\end{proof}

\begin{lemma}[Averaged estimation error with compression]
\label{dmx:lem:lr-estimation}
Under the assumptions and algorithm of Theorem~\ref{dmx:thm:lr},
\begin{equation}
\frac1T\sum_{t=0}^{T-1}\mathbb E\|\bar M_t-H_t\|_F
\le\sqrt{\mathcal V_T}+R_T.
\label{dmx:eq:lr-estimation}
\end{equation}
\end{lemma}
\begin{proof}
Let $\mathcal F_t$ denote the history immediately before drawing the
fresh samples at step $t+1$, after any model averaging. Both $W_t^{(k)}$
and $W_{t+1}^{(k)}$ are $\mathcal F_t$-measurable. Define
\begin{align*}
\zeta_{t+1}^{(k)}={}&\nabla f^{(k)}(W_{t+1}^{(k)};\xi_{t+1}^{(k)})
                    -\nabla f^{(k)}(W_{t+1}^{(k)}),\\
\nu_{t+1}^{(k)}={}&\nabla f^{(k)}(W_{t+1}^{(k)};\xi_{t+1}^{(k)})
                  -\nabla f^{(k)}(W_t^{(k)};\xi_{t+1}^{(k)})\\
&-\bigl(\nabla f^{(k)}(W_{t+1}^{(k)})-\nabla f^{(k)}(W_t^{(k)})\bigr),\\
Z_{t+1}={}&\frac1K\sum_k\bigl(a\zeta_{t+1}^{(k)}+\beta\nu_{t+1}^{(k)}\bigr).
\end{align*}
The same-sample construction and conditional independence give
\begin{equation}
\mathbb E[Z_{t+1}\mid\mathcal F_t]=0,\qquad
\mathbb E[\|Z_{t+1}\|_F^2\mid\mathcal F_t]
\le\frac{2a^2\sigma^2}{K}
 +\frac{2L^2}{K^2}\sum_k\|W_{t+1}^{(k)}-W_t^{(k)}\|_F^2.
\label{dmx:eq:innovation}
\end{equation}
In particular, no independence is assumed between the two terms at the
same node. The estimate uses $\|A+B\|_F^2\le2\|A\|_F^2+2\|B\|_F^2$,
centering of the stochastic gradient difference, and $\beta^2\le1$.

For $e_t=\bar M_t-H_t$, the actual recursion is
\begin{equation}
e_{t+1}=\beta e_t+Z_{t+1}+E_{t+1}^{\mathrm{comp}}.
\label{dmx:eq:compressed-recursion}
\end{equation}
The residual $E_{t+1}^{\mathrm{comp}}$ is generally biased and depends on the fresh
samples, so it cannot be treated as a centered innovation. Instead,
define two auxiliary processes on the actual LR trajectory:
\[
Q_0=e_0,\quad B_0^{\mathrm{comp}}=0,\qquad
Q_{t+1}=\beta Q_t+Z_{t+1},\quad
B_{t+1}^{\mathrm{comp}}=\beta B_t^{\mathrm{comp}}+E_{t+1}^{\mathrm{comp}}.
\]
Induction gives $e_t=Q_t+B_t^{\mathrm{comp}}$.
The process $Q_t$ is measurable with respect to $\mathcal F_t$;
it is an analytical decomposition, not an estimator on a separate
uncompressed run. Thus \eqref{dmx:eq:innovation} implies
\[
\mathbb E\|Q_{t+1}\|_F^2
\le\beta^2\mathbb E\|Q_t\|_F^2
 +\frac{2a^2\sigma^2}{K}
 +\frac{2L^2}{K^2}\sum_k\mathbb E\|W_{t+1}^{(k)}-W_t^{(k)}\|_F^2.
\]
The dense initialization gives $\mathbb E\|Q_0\|_F^2\le\sigma^2/(KB_0)$.
Unroll the recursion, sum its nonnegative geometric coefficients using
$(1-\beta^2)^{-1}\le a^{-1}$, and apply
Lemma~\ref{dmx:lem:lr-consensus}. This gives explicitly
\begin{align*}
\frac1T\sum_{t=0}^{T-1}\mathbb E\|Q_t\|_F^2
&\le\frac{\sigma^2}{KB_0aT}
 +\frac{2a\sigma^2}{K}
 +\frac{2L^2}{aK^2T}\sum_{t=0}^{T-1}\sum_k
  \mathbb E\|W_{t+1}^{(k)}-W_t^{(k)}\|_F^2\\
&\le\frac{\sigma^2}{KB_0aT}
 +\frac{2a\sigma^2}{K}
 +\frac{2L^2\eta^2d(4\tau-3)}{aK}
 =\mathcal V_T.
\end{align*}
Jensen's inequality bounds the average expected norm of $Q_t$ by
$\sqrt{\mathcal V_T}$.

For the second process, unrolling gives
$B_t^{\mathrm{comp}}=\sum_{s=1}^t\beta^{t-s}E_s^{\mathrm{comp}}$. Hence, by the triangle
inequality and finite summation,
\begin{align*}
\frac1T\sum_{t=0}^{T-1}\mathbb E\|B_t^{\mathrm{comp}}\|_F
&\le\frac1T\sum_{s=1}^{T-1}\mathbb E\|E_s^{\mathrm{comp}}\|_F
                 \sum_{t=s}^{T-1}\beta^{t-s}\\
&=\frac1{aT}\sum_{s=1}^{T-1}(1-\beta^{T-s})\mathbb E\|E_s^{\mathrm{comp}}\|_F
 \le R_T.
\end{align*}
Combining the two bounds proves \eqref{dmx:eq:lr-estimation}.

All moments used above are finite. Bounded directions bound the model
iterates on each finite horizon. Smoothness and bounded variance give
finite second moments of the sampled gradients. Finally,
the truncated-SVD inequality $\|A_r\|_F\le\|A\|_F$ and the triangle inequality for
averaging propagate finite second moments through the LR recursion.
No interchange of infinite sums and expectations is needed.
\end{proof}

\begin{proof}[Proof of Theorem~\ref{dmx:thm:lr}]
Lemma~\ref{dmf:lem:descent} depends on exact polar alignment, objective
smoothness, and mean-preserving model averaging. It does not require
mean-preserving momentum synchronization. Those three conditions hold
on the actual LR trajectory, so the same descent inequality applies.
By Lemma~\ref{dmx:lem:lr-consensus}, its model and momentum disagreement
terms have the same bounds as before. Replacing the averaged estimation
bound by Lemma~\ref{dmx:lem:lr-estimation} and telescoping gives
\begin{align*}
\frac1T\sum_{t=0}^{T-1}\mathbb E\|\nabla f(\bar W_t)\|_F
\le{}&\frac{\Delta}{\eta T}+\frac{\eta Ld}{2}+4\eta Ld h
 +4\beta\eta Ld h+8a\eta Ld h^2\\
&+2\sqrt d\,ah(\delta+2\sigma)
 +2\sqrt d\bigl(\sqrt{\mathcal V_T}+R_T\bigr).
\end{align*}
The right-hand side equals
$B_T^{\mathrm{ex}}+2\sqrt d\,R_T$, where $B_T^{\mathrm{ex}}$ denotes the right-hand side of \eqref{dmf:eq:main-bound}.
There are $N_T=\lfloor(T-1)/\tau\rfloor$
contributing communication boundaries. Thus,
\[
R_T
\le \frac{N_T\rho_{r,T}}{aT}
\le \frac{\rho_{r,T}}{a\tau},
\]
where $a=1-\beta$. Consequently,
\[
\frac{1}{T}\sum_{t=0}^{T-1}
\mathbb{E}\|\nabla f(\bar W_t)\|_F
\le B_T^{\mathrm{ex}}
+\frac{2\sqrt d\,\rho_{r,T}}{a\tau}.
\]
Under the parameter choices of
Theorem~\ref{dmf:thm:main},
$B_T^{\mathrm{ex}}\le C_\tau(KT)^{-1/3}$
and $a=K^{1/3}/(2T^{2/3})$. Substitution yields
\[
\frac{1}{T}\sum_{t=0}^{T-1}
\mathbb{E}\|\nabla f(\bar W_t)\|_F
\le
\frac{C_\tau}{(KT)^{1/3}}
+\frac{4\sqrt d\,T^{2/3}}{\tau K^{1/3}}\rho_{r,T},
\]
which proves Theorem~\ref{dmx:thm:lr}.
If $N_T=0$, then $R_T=\rho_{r,T}=0$.
\end{proof}

\begin{proof}[Proof of Corollary~\ref{dmx:cor:hybrid}]
Lemma~\ref{dmx:lem:ns} bounds the operator and Frobenius norms of the
NS directions by one and $\sqrt d$, respectively, on the actual LR--NS
trajectory. The model-displacement and momentum-consensus arguments in
Lemma~\ref{dmx:lem:lr-consensus} therefore still apply. The operator-norm
model disagreement is at most $2\eta h$, as in
\eqref{dmx:eq:ns-model}. The decomposition in
Lemma~\ref{dmx:lem:lr-estimation} uses only the sampling assumptions,
STORM recursion, direction bounds, and two-sided compression residual,
so it gives the same $\sqrt{\mathcal V_T}+R_T$ estimate.
Finally, use the NS descent inequality \eqref{dmx:eq:ns-descent},
telescope, and divide by $\eta T(1-\varepsilon_q)$. The resulting bound is
the bound in Corollary~\ref{dmx:cor:hybrid}, with no additional decay condition
on the compression residuals.
\end{proof}

\section{Additional Experimental Details and Results}
\label{app:exp_details}

This section provides the implementation details and additional analysis of the experimental
results in Section~\ref{sec:experiments}.
All experiments run on a single NVIDIA A100-SXM4
GPU with 40\,GB memory, with the five nodes simulated
sequentially. We use FP32 computation and disable TF32.
\subsection{Additional Details for CIFAR-10}
\label{app:cifar_details}
\textbf{Training protocol.}
We split the original training set into 45,000 training
and 5,000 validation examples, and distribute the
training examples across five fully participating
nodes in a class-balanced IID manner.
Each node performs $\tau=5$ local updates per
communication round with a batch size of 128.
We train for 80 sample-equivalent epochs using
seeds 2027, 2028, and 2029, and report the mean
and sample standard deviation of final-model
test performance. Hyperparameters are selected
using validation loss.

\textbf{Network architecture.}
We use a widened ResNet-20 with an initial $3\times3$
convolution and three residual stages of widths
64, 128, and 256. Each stage contains three residual
blocks, each with two $3\times3$ convolutions.
The network uses four-group Group Normalization
and ends with global average pooling and a linear
classifier.

\textbf{Matrix parameters.}
For the 18 hidden convolutional layers, we represent
each kernel
$W\in\mathbb{R}^{c_{\mathrm{out}}\times c_{\mathrm{in}}
\times k_h\times k_w}$
as a matrix
$\widetilde{W}\in\mathbb{R}^{c_{\mathrm{out}}\times
(c_{\mathrm{in}}k_hk_w)}$.
Our spectral updates operate on this matrix representation,
which is reshaped back for convolution.
The input convolution, normalization parameters,
and classification head are updated using AdamW.

\textbf{Newton--Schulz approximation.}
Given an input matrix $M$, we initialize
$X_0=M/\|M\|_F$, with zero inputs mapped to zero.
For computational efficiency, we transpose $M$
when it has more rows than columns and transpose
the output back afterward.
We apply the practical quintic iteration
\[
X_{j+1}
= aX_j+b(X_jX_j^\top)X_j
+c(X_jX_j^\top)^2X_j,
\]
where $(a,b,c)=(3.4445,-4.775,2.0315)$.
We use $q=5$ iterations for training and evaluate
$q\in\{3,5,8,10\}$ in the operator study.
This practical quintic iteration is distinct from
the cubic iteration analyzed theoretically.

For the same input $M=U\Sigma V^\top$, let
$D_{\mathrm{SVD}}=U_rV_r^\top$ denote the reference
polar direction on its numerical singular-value
support, and let $D_{\mathrm{NS},q}$ denote the
NS output after $q$ iterations.
We measure their relative Frobenius discrepancy as
\[
\varepsilon_q =
\frac{\|D_{\mathrm{NS},q}-D_{\mathrm{SVD}}\|_F}
     {\|D_{\mathrm{SVD}}\|_F}.
\]
This metric captures both directional and magnitude
differences (Figure~\ref{fig:q}).

\textbf{Learning rates.}
The peak learning rate is $1$ for FedAvg and $0.03$
for FedAvg-SGDM, FedMuon-FA, and all \algn{} variants.
The auxiliary AdamW optimizer uses a peak learning
rate of $10^{-3}$.
Learning rates follow a five-epoch linear warmup
and cosine decay to $1\%$ of their peak values.

\begin{table}[!htb]
\centering
\caption{Final test accuracy on CIFAR-10 after 80 epochs.
Results are mean $\pm$ sample standard deviation
over three random seeds.}
\label{tab:test_accuracy}
\small
\begin{tabular}{lc}
\toprule
Method & Test accuracy (\%) \\
\midrule
FedAvg                       & $84.21 \pm 1.46$ \\
FedAvg-SGDM                         & $86.37 \pm 0.24$ \\
FedMuon-FA                   & $86.95 \pm 0.11$ \\
FedMuon-SF                & $87.31 \pm 0.32$ \\
DisMUSE-SVD                  & $89.69 \pm 0.10$ \\
DisMUSE-NS                   & $\mathbf{89.92 \pm 0.07}$ \\
DisMUSE-LR ($r=16$)          & $80.52 \pm 0.88$ \\
DisMUSE-Hybrid ($r=16$)      & $81.76 \pm 0.26$ \\
DisMUSE-LR ($r=32$)          & $85.08 \pm 0.64$ \\
DisMUSE-Hybrid ($r=32$)      & $85.03 \pm 0.43$ \\
\bottomrule
\end{tabular}

\end{table}

\subsection{Additional Details for Tiny Shakespeare}
\label{app:fashion_details}

\textbf{Training protocol.}
We split the corpus into contiguous training, validation,
and reserved evaluation segments containing 80\%, 10\%,
and 10\% of the characters, respectively.
The vocabulary is fitted on the training segment, with an
additional unknown-character token.
Each node independently samples batches of 32 sequences
of length 256 from the shared training pool.
The full protocol uses 1,500 local updates per node and
seeds 3101, 3102, and 3103.
We evaluate token-averaged validation cross-entropy and
perplexity every 100 updates and compare final-model metrics.

\textbf{Network architecture.}
The four-layer causal Transformer uses hidden dimension 256,
eight attention heads, feed-forward dimension 1,024,
GELU activation, post-layer normalization, and dropout 0.1.
We use sinusoidal positional encoding and an additional
layer normalization before the output projection.
Input embeddings and output weights are not tied.

\textbf{Matrix parameters.}
Spectral updates apply to the attention output projection
and both feed-forward weight matrices in each layer,
with dimensions $256\times256$, $1024\times256$, and
$256\times1024$, respectively.
The remaining parameters, including attention QKV weights,
embeddings, normalization parameters, output-head parameters,
and biases, use AdamW.
Paired gradient evaluations in \alg share both the
sampled batch and dropout randomness.

\textbf{Learning rates.}
FedAdam uses peak learning rate 0.003 and moment
coefficients $(0.9,0.95)$; its optimizer states remain local
and persist across aggregation rounds.
FedAvg uses SGD without momentum and peak learning rate 0.1.
\alg uses matrix learning rate 0.03, estimator coefficient
$\beta=0.9$, and an initialization batch of 128 sequences
per node.
Auxiliary AdamW uses learning rate $10^{-3}$ and moment
coefficients $(0.9,0.999)$.
The weight-decay coefficient is $5\times10^{-4}$.
Warmup occupies the first 5\% of updates, followed by cosine
decay to 1\% of the peak learning rate.
For NS, we use $q=5$ quintic iterations with coefficients
$(a,b,c)=(3.4445,-4.7750,2.0315)$ and Frobenius-norm
normalization of the input matrix.

\begin{table}[!htb]
    \centering
    \caption{Validation perplexity on Tiny Shakespeare after
    1500 training steps. We report the mean and sample standard
    deviation over three random seeds. Lower is better.}
    \label{tab:shakespeare_ppl}
    \begin{tabular}{lc}
        \toprule
        Method & Perplexity $\downarrow$ \\
        \midrule
        FedAvg & $12.254 \pm 0.014$ \\
        FedAdam & $5.789 \pm 0.069$ \\
        \midrule
        DisMUSE-SVD & $\mathbf{4.293 \pm 0.010}$ \\
        DisMUSE-NS & $4.320 \pm 0.019$ \\
        DisMUSE-LR-SVD ($r=32$) & $4.651 \pm 0.020$ \\
        DisMUSE-LR-NS ($r=32$) & $4.707 \pm 0.006$ \\
        DisMUSE-LR-SVD ($r=64$) & $4.487 \pm 0.011$ \\
        DisMUSE-LR-NS ($r=64$) & $4.535 \pm 0.010$ \\
        \bottomrule
    \end{tabular}
\end{table}
Table~\ref{tab:shakespeare_ppl} reports the final validation
perplexity on Tiny Shakespeare over three random seeds.
\alg-NS closely matches exact-SVD \alg, with mean
perplexities of 4.320 and 4.293, respectively.
The low-rank variants also outperform FedAvg and FedAdam
under this experimental setting, while increasing the rank
from 32 to 64 consistently improves perplexity for both
the NS and SVD variants.

\end{document}